%% file: main_arxiv.tex
\documentclass{article}
\PassOptionsToPackage{numbers, compress}{natbib}
\usepackage[preprint]{neurips_2026}

\usepackage[utf8]{inputenc}
\usepackage[T1]{fontenc}
\usepackage{amsfonts}
\usepackage{amsmath}
\usepackage{amssymb}
\usepackage{booktabs}
\usepackage{graphicx}
\usepackage{listings}
\usepackage{microtype}
\usepackage{nicefrac}
\usepackage{url}
\usepackage{xcolor}
\usepackage{colortbl}
\usepackage{xspace}
\usepackage{subcaption}
\usepackage{wrapfig}
\usepackage{titlesec}

\usepackage{hyperref}
\usepackage[capitalize]{cleveref}

\definecolor{codekw}{RGB}{0,74,173}
\definecolor{codestr}{RGB}{163,21,21}
\definecolor{codecom}{RGB}{0,128,96}
\definecolor{codeid}{RGB}{92,38,153}
\definecolor{codenum}{RGB}{128,64,0}
\definecolor{Maroon}{RGB}{128,0,0}
\newcommand{\best}[1]{\textcolor{Maroon}{\textbf{#1}}}

\lstdefinestyle{pythoncode}{
  language=Python,
  basicstyle=\ttfamily\fontsize{5}{5.2}\selectfont,
  columns=fullflexible,
  keepspaces=true,
  breaklines=true,
  breakatwhitespace=false,
  showstringspaces=false,
  keywordstyle=\bfseries\color{codekw},
  commentstyle=\itshape\color{codecom},
  stringstyle=\color{codestr},
  identifierstyle=\color{codeid},
  numberstyle=\color{codenum},
  upquote=true,
  frame=single,
  framerule=0.2pt,
  rulecolor=\color{black!15},
  framesep=2pt,
  literate=
    {0}{{{\color{codenum}0}}}{1}
    {1}{{{\color{codenum}1}}}{1}
    {2}{{{\color{codenum}2}}}{1}
    {3}{{{\color{codenum}3}}}{1}
    {4}{{{\color{codenum}4}}}{1}
    {5}{{{\color{codenum}5}}}{1}
    {6}{{{\color{codenum}6}}}{1}
    {7}{{{\color{codenum}7}}}{1}
    {8}{{{\color{codenum}8}}}{1}
    {9}{{{\color{codenum}9}}}{1},
  xleftmargin=0pt,
  xrightmargin=0pt,
  aboveskip=0pt,
  belowskip=0pt
}

\newcommand{\agentname}{Articraft\xspace}
\newcommand{\datasetname}{Articraft-10K\xspace}
\newcommand{\eg}{e.g.\xspace}

\newcommand{\toolyes}{\textcolor{green!50!black}{$\checkmark$}}
\newcommand{\toolno}{\textcolor{red!70!black}{$\times$}}

\newcommand\rurl[1]{%
  \href{https://#1}{\nolinkurl{#1}}%
}

\makeatletter
\renewcommand{\paragraph}{%
  \@startsection{paragraph}{4}%
  {\z@}{-0em}{-0.5em}%
  {\normalfont\normalsize\bfseries}%
}
\makeatother

\titlespacing*{\section}
{0pt}{0.6\baselineskip}{0.3\baselineskip}

\titlespacing*{\subsection}
{0pt}{0.4\baselineskip}{0.2\baselineskip}

\title{\agentname: An Agentic System for \\ Scalable Articulated 3D Asset Generation}

\author{
Matt Zhou$^{1\#*}$ \quad
Ruining Li$^{2*}$ \quad
Xiaoyang Lyu$^{1*}$ \quad
Zhaomou Song$^{1*}$ \quad
Zhening Huang$^{1*}$ \\[0.2em]
\textbf{Chuanxia Zheng}$^{3}$ \quad
\textbf{Christian Rupprecht}$^{2}$ \quad
\textbf{Andrea Vedaldi}$^{2}$ \quad
\textbf{Shangzhe Wu}$^{1}$ \\[0.5em]
$^{1}$University of Cambridge \quad
$^{2}$University of Oxford  \quad
$^{3}$Nanyang Technological University \\[0.2em]
\small\rurl{articraft3d.github.io}
}

\begin{document}
\maketitle

\begingroup
\renewcommand{\thefootnote}{}
\footnotetext{$\#$ Project lead. \quad $*$ Core technical contribution.}
\endgroup

\begin{figure}[h]
\vspace{-0.4in}
\centering
\includegraphics[trim={0 15bp 0bp 0}, clip, width=\linewidth]{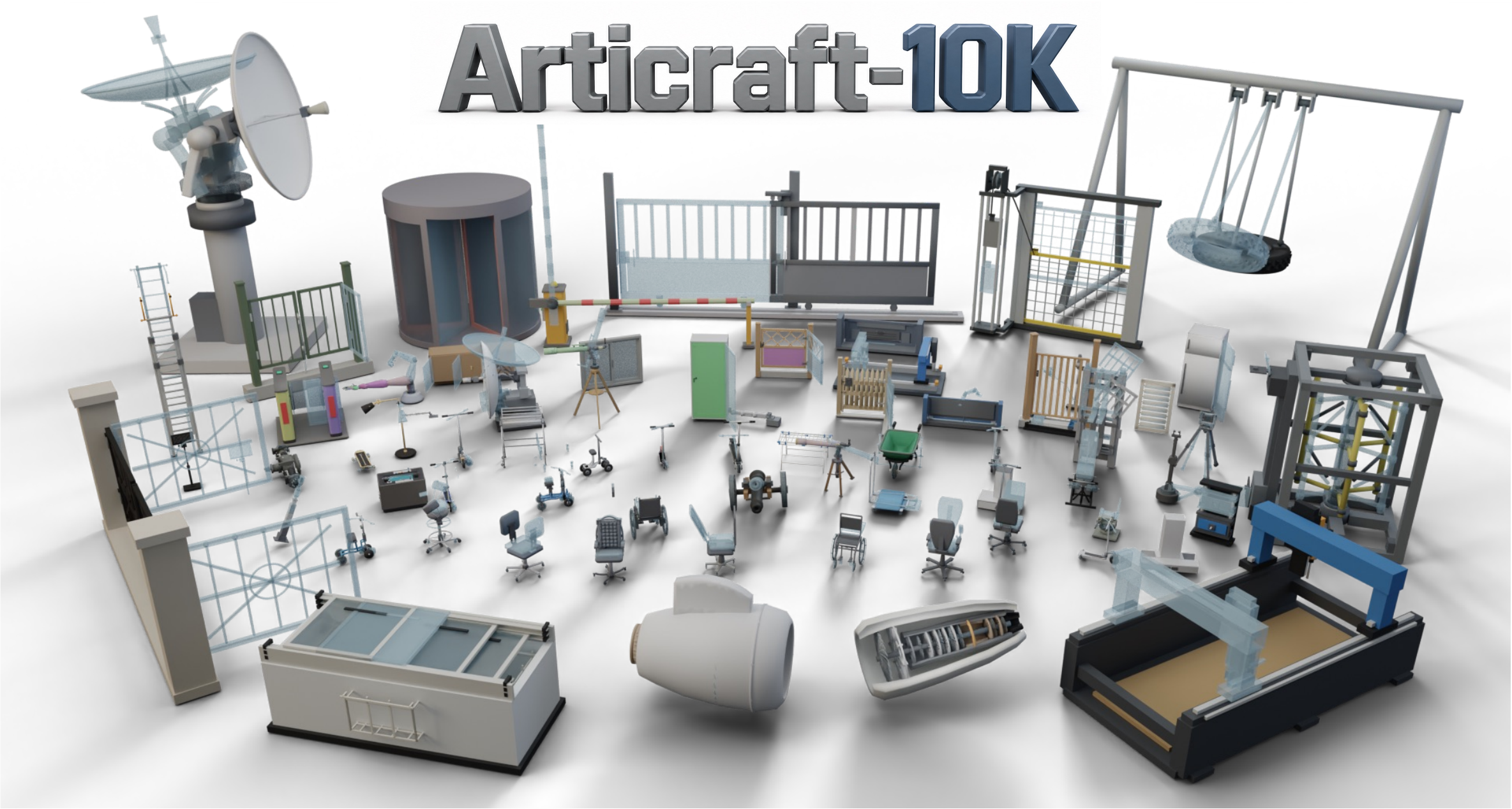}
\caption{We introduce \agentname{}, an agentic system for scalable articulated 3D asset generation.
Using \agentname{}, we created \datasetname{}, a large-scale dataset of over 10K articulated 3D assets spanning a wide range of everyday object categories.
}%
\vspace{-0.1in}
\label{fig:teaser}
\end{figure}

\input{sec/0_abstract}
\input{sec/1_intro}
\input{sec/2_related}
\input{sec/3_method}

\input{sec/4_dataset}

\input{sec/5_experiments}
\input{sec/45_dataset_experiments}

\input{sec/6_conclusions}
\input{sec/7_acknowledgements}

{
\small
\raggedright
\setlength{\emergencystretch}{3em}
\bibliographystyle{ieeenat_fullname}
\bibliography{ref,vedaldi_general,vedaldi_specific}
}

\newpage
\appendix

\input{appendices/appendix1}

\end{document}

%% file: sec/0_abstract.tex
\begin{abstract}
A bottleneck in learning to understand articulated 3D objects is the lack of large and diverse datasets.
In this paper, we propose to leverage large language models (LLMs) to close this gap and generate articulated assets at scale.
We reduce the problem of generating an articulated 3D asset to that of writing a program that builds it.
We then introduce a new agentic system, \agentname, that writes such programs automatically.
We design a programmatic interface and harness to help the LLM do so effectively.
The LLM writes code against a domain-specific SDK for defining parts, composing geometry, specifying joints, and writing tests to validate the resulting assets.
The harness exposes a restricted workspace and interface to the LLM, validates the resulting assets, and returns structured feedback.
In this way, the LLM is not distracted by details such as authoring a URDF file or managing a complex software environment.
We show that this produces higher-quality assets than both state-of-the-art articulated-asset generators and general-purpose coding agents.
Using \agentname{}, we build \datasetname{}, a curated dataset of over 10K articulated assets spanning 245 categories, and show its utility both for training models of articulated assets and in downstream applications such as robotics simulation and virtual reality.
\end{abstract}

%% file: sec/1_intro.tex
\section{Introduction}%
\label{sec:introduction}

Articulated objects, such as cabinets, pliers, desk lamps, strollers, folding chairs, and industrial tools, are ubiquitous in the real world and thus important to applications in 3D content creation, gaming, physical design, and robotics.
Because their function depends on the motion of their parts as much as on their geometry, modelling shape alone is insufficient: a useful representation must also capture part hierarchy, joints, and range of motion.

A central obstacle in this area is data.
Existing datasets of articulated assets are small, narrow in coverage, and uneven in quality~\citep{liu2025articulatedsurvey,xiang20sapien:,wang19shape2motion:,liu22akb-48:,geng23gapartnet:,wang24grutopia:}.
In particular, many object categories of practical importance are underrepresented.
As a result, learning-based methods overfit the available data and fail to generalise to new categories~\cite{li2026particulate}.

One way to close this gap is to \emph{generate} articulated assets synthetically.
This is itself difficult, but recent agentic systems suggest a path forward.
Tools such as Claude Code~\cite{anthropic2026claudecode} and Codex~\cite{openai2026codexcli} can plan, write, debug, and iteratively refine software with little human supervision.
We hypothesise that these advances transfer to 3D generation, because articulated objects share the recursive, compositional structure of programs.
Designing an object is a sequence of decisions: decompose the object into parts, decide how the parts connect, specify joints and motion limits, instantiate geometry, validate behaviour, and revise on failure.
This workflow resembles agentic programming far more than one-shot generation.

To test this hypothesis, we introduce \emph{\agentname}, an agent that builds articulated 3D objects at scale by generating code.
We design \agentname around three principles.
It should be \emph{automatic}: given an object description, it produces an articulated asset without manual intervention.
It should be \emph{lightweight}: to keep per-asset inference cost low and make large-scale data generation feasible, it avoids heavy external graphics software (\eg, Blender) and image-based feedback.
It should be \emph{expressive}: it covers a wide range of articulated categories, including complex mechanisms.

Our key technical innovation is the design of the agent, which rests on two components: an \emph{agent harness} and an LLM-friendly \emph{Software Development Kit} (SDK).
Together, they enable off-the-shelf large language models (LLMs) to generate articulated 3D assets without retraining, drawing on the LLM's coding ability and its prior knowledge of how everyday objects are structured and move.
This design distinguishes our approach from prior LLM-based methods for articulated asset generation~\cite{xie25text-to-cadquery:,guan2025cadcoder,le25articulate-anything:}.

The SDK makes it easier for the LLM to understand how to generate assets.
The LLM is asked to write a single program against this SDK\@; once executed, the program produces a complete articulated asset.
The SDK is focused, expressive, and LLM-friendly: it spans a wide range of articulated categories while staying close enough to familiar coding patterns that generation remains reliable.
It exposes both low-level primitives (\eg, adding a cylinder) and high-level abstractions (\eg, adding a hinge), keeping programs compact and readable, and it lets the LLM write and execute object-specific validation code to test structural integrity.

The harness turns the LLM into an iterative agent rather than a single-pass generator.
The harness exposes a minimal workspace and interface: edit a single program, execute it, and receive or request feedback on the resulting 3D asset.
Its design encodes deliberate choices about how much context to expose, how that context is structured, how generated assets are validated, and when to stop iterating.

Together, the focused SDK and minimal harness keep per-asset inference cost low, which is what makes large-scale generation practical. This design produces realistic articulated objects across a diverse set of categories.
It outperforms both prior work in this area and general-purpose coding agents such as Codex~\cite{openai2026gpt53codex} and Claude Code~\cite{anthropic2026claudecode}.
It is also substantially more lightweight, avoiding the heavy external graphics software (\eg, Blender) that prior approaches often rely on.
Using \agentname, we build \emph{\datasetname}, a curated dataset of over 10K articulated objects spanning 245 categories.
As one application, we retrain \emph{Particulate}~\cite{li2026particulate}, a model that estimates the articulated structure of 3D objects, and obtain a substantial performance boost.
We also showcase applications in robotics simulation and virtual reality.

We will release \datasetname publicly, together with the code representations of all assets, the agent's reasoning traces, and the agent environment itself, which can be used with different LLM backends.

%% file: sec/2_related.tex
\section{Related Work}%
\label{sec:related}

\input{tables/dataset_comparison}

\paragraph{Generating articulated objects.}

Recent work has begun to generate articulated 3D assets directly.
NAP~\cite{lei23nap:} and CAGE~\cite{liu24cage:} propose ad-hoc representations of articulated objects that can be generated using denoising diffusion~\cite{song21denoising}.
ArtFormer~\cite{su25artformer:} and MeshArt~\cite{gao25meshart:} do so with a token-based representation and an auto-regressive generative transformer.
URDFormer~\cite{chen2024urdformer}, URDF-Anything~\cite{li25urdf-anything:} and URDF-Anything+~\cite{wu26urdf-anything:} learn to map images to URDF proxy representations, the latter
building on AutoPartGen~\cite{chen25autopartgen} to extract a URDF object sequentially, one part at a time.
Real2Code~\cite{zhao2025real2code} starts by reconstructing an image in 3D and performing part segmentation and then uses LLM-based code generation to infer the articulation parameters.
SINGAPO~\cite{liu25singapo:} generates articulated object parts from a single image.
PhysX-3D~\cite{cao25physx-3d:} augments the structured 3D latents~\cite{xiang25structured} to generate physical attributes in addition to 3D shape and PhysX-Anything~\cite{cao26physx-anything:} predicts simulation-ready geometry, articulation, and physical attributes from a single image.
More similar to our approach is Articulate-Anything~\cite{le25articulate-anything:}, which uses VLM agents to generate Python code to construct articulated URDF assets from text, images, or videos.
However, they rely on retrieving part meshes from an existing 3D asset library, which constrains category diversity.
A more recent work, ArtiCAD~\cite{shui2026articad}, generates articulated CAD assemblies with a multi-agent pipeline, but relies on visual feedback from multi-view renders and joint-motion keyframes during review.
In contrast, \agentname{} is entirely code-based, avoiding image-based feedback and keeping large-scale generation lightweight.

\paragraph{Programmatic and agentic 3D design.}

Several prior works generate 3D objects by controlling CAD software.
ShapeAssembly~\cite{jones20shapeassembly:} introduces an SDK to simplify interfacing with CAD software and target it with an ad-hoc code generator.
ShapeMOD~\cite{jones21shapemod:} further discovers reusable programmatic primitives in this space.
DeepCAD~\cite{wu21deepcad:} and Text2CAD~\cite{khan24text2cad:} generate suitably-encoded CAD commands using an autoregressive transformer.
Text-to-CadQuery~\cite{xie25text-to-cadquery:} and CAD-Coder~\cite{guan2025cadcoder} write Python code against the CadQuery API\@.
Our agent targets a superset of these APIs and produces articulated, simulation-ready assets instead of static ones.

Our method is also related to language agents that interleave reasoning, action, tool use, and feedback, especially ReAct~\cite{yao23react:}, Reflexion~\cite{shinn23reflexion:}, SWE-agent~\cite{xia2024sweagent}, and code-as-action methods such as PAL~\cite{gao23pal:} and Code as Policies~\cite{liang23code}.
SWE-agent, in particular, shows that LLMs benefit from using software interfaces that are specially designed for the task and that provide feedback in a well-structured format.
We follow the same principle, including in the form of a domain-specific SDK that the agent can use to write a program.

\paragraph{Articulated assets and simulation datasets.}

PartNet-Mobility~\cite{xiang20sapien:} is perhaps the most widely-used dataset of articulated objects to date based on synthetic assets.
AKB-48~\cite{liu22akb-48:} starts instead from scans of real objects, and GAPartNet~\cite{geng23gapartnet:} extends PartNet with annotations for part-based affordances.
GRScenes, introduced as part of GRUtopia~\cite{wang24grutopia:}, generates entire scenes instead of single objects.
RoboCasa365~\cite{nasiriany26robocasa365:} provides simulated environments for robotics with more than a hundred object categories, of which only a dozen are articulated.
Infinigen-Sim~\cite{joshi25infinigen-sim:} follows Infinigen~\cite{raistrick23infinite} and manually defines procedural generators to generate articulated 3D assets instead of scenes.
PhysX-3D~\cite{cao25physx-3d:} introduces PhysXNet, a collection of articulated 3D objects created semi-automatically.
We use our new agent to create \datasetname, a large collection of curated articulated assets with substantially broader category coverage than these prior works (\cref{tab:dataset_comparison}).

%% file: tables/dataset_comparison.tex
\begin{table}[t]
\centering
\caption{Comparison between \datasetname and existing articulated or physics-annotated 3D object datasets. For datasets that also contain non-articulated assets, the counts refer to the portion of articulated or interactive objects, where reported.}%
\label{tab:dataset_comparison}
\footnotesize
\setlength{\tabcolsep}{4pt}
\renewcommand{\arraystretch}{1.08}
\begin{tabular}{lccl}
\toprule
\textbf{Dataset} & \textbf{\# Categories} & \textbf{\# Assets} & \textbf{Source of 3D Geometry} \\
\midrule
PartNet-Mobility~\citep{xiang20sapien:} & $46$ & $2.3$K & PartNet~\citep{mo19partnet:} \\
AKB-48~\citep{liu22akb-48:} & $48$ & $2.0$K & Real scanning \\
GAPartNet~\citep{geng23gapartnet:} & $27$ & $1.2$K & PartNet-Mobility~\citep{xiang20sapien:}, AKB-48~\citep{liu22akb-48:} \\
GRScenes~\citep{wang24grutopia:} & $22$ & $1.8$K & Human artist design \\
Infinigen-Sim~\citep{joshi25infinigen-sim:} & $18$ & $20$K & Procedural generation \\
PhysXNet~\citep{cao25physx-3d:} & $24$ & $26$K & PartNet~\citep{mo19partnet:} \\
PhysXNet-XL~\citep{cao25physx-3d:} & $11$ & $6$M & Procedural generation \\
PhysX-Mobility~\citep{cao26physx-anything:} & $47$ & $2$K & PartNet-Mobility~\citep{xiang20sapien:} \\
RoboCasa365~\citep{nasiriany26robocasa365:} & $12$ & $0.5$K & Human artist design \\
\midrule
\textbf{\datasetname{} (Ours)} & $245$ & $\!>\!10$K & Agentic generation \\
\bottomrule
\end{tabular}
\vspace{-1.8em}
\end{table}

%% file: sec/3_method.tex
\section{\agentname}%
\label{sec:method}

\input{figures/fig_method}

We introduce \agentname, an agent that writes programs to build articulated 3D assets.
Given a natural-language description $x$ and, optionally, a reference image of the object, \agentname writes a Python program $y$ which, once executed, outputs an articulated 3D asset $a$.
The asset consists of a URDF\footnote{Unified Robot Description Format, a commonly used XML format for representing articulated 3D assets.} containing
3D meshes, semantic parts, and articulated joints with their axes and motion ranges.

\agentname builds 3D assets iteratively, following the reasoning-action-observation pattern of language agents~\citep{yao23react:,shinn23reflexion:,wang25voyager:}.
The agent is an off-the-shelf LLM $E$ with coding capabilities.
The LLM is given a system prompt $p$ (see \cref{sec:appendix_details}) that specifies the object generation task and the programmatic interface and harness that must be used to solve the task.
A harness exposes to the LLM a workspace and interface to manipulate the program.
The workspace maintains a state consisting of the current program $y_t$, the current asset $a_t$, as well as a history $h_t$ of past revisions.
The LLM $E(p, x, h_t)$ takes the system prompt $p$, the user prompt $x$, and the current history $h_t$ and outputs one or more commands for the harness.
These commands act on the current state $(y_t, a_t)$ by editing the program, compiling it into a new asset, or probing the asset to obtain feedback.
Thus the harness $C$ outputs a new state $(y_{t+1}, a_{t+1})$ and feedback $s_{t+1}$, which are appended to the history for the next iteration:
\vspace{-0.05em}
$$
(y_{t+1}, a_{t+1}, s_{t+1}) = C(E(p,x, h_t),y_t, a_t),
\quad
h_{t+1} = h_t \cup \{(y_{t+1}, a_{t+1}, s_{t+1})\}.
$$
\vspace{-0.05em}%
The process terminates when the LLM does not issue further editing commands and when the validation criteria are met.
The last version of the asset $a_T$ is accepted as final output.

A key innovation of \agentname is to provide the LLM backend with 
a programmatic interface (\cref{sec:object_programs}) 
and a harness (\cref{sec:agentic_authoring_interface})
specific to articulated 3D design (\cref{fig:method_overview}).
This follows the principle that agent-computer interfaces should be task-specific instead of generic~\citep{xia2024sweagent} and significantly improves the effectiveness of the backend, which is otherwise a general purpose off-the-shelf coding model.
In this manner, \agentname does not need to use \emph{visual} feedback used in prior work~\cite{le25articulate-anything:}, which is expensive to produce and use.
Instead, the harness and programmatic interface provide specialised tools for authoring and checking the geometry of the object directly and efficiently.

\subsection{The \agentname Programmatic Interface}%
\label{sec:object_programs}

\input{figures/sample_program}

Each 3D asset $a$ generated by \agentname is defined by a program $y$ consisting of a single Python file, \verb!model.py! (see the example in \cref{fig:sample_program}).
The file exposes two entry points:
\verb!build_object_model()! to construct the articulated object, and
\verb!run_tests()! to record prompt-specific geometric checks and explicit decisions to relax some of these.
It then binds the constructed object to the Python variable \verb!object_model!, which the harness can use to extract the generated asset.
An implicit \emph{contract} separates object authoring from details of system execution, such as file management, URDF export, and validation.
The agent only needs to write code against the SDK given below to specify the object parts, their geometry and articulation, and what object-specific tests should hold, and the harness handles the rest.
This keeps the editable target expressive but small.

\paragraph{Defining Geometry and Parts.}%
\label{sec:part_geometry_representation}

The program \verb!model.py! uses the SDK to build the object.
At the top level, the program constructs an \verb!ArticulatedObject! and populates it with named \verb!Part!s, which provide the semantic scaffold for the asset.
Parts can be connected by joints and named as targets for tests and geometric probes.
The SDK also contains a variety of tools for defining the geometry of the parts, from low-level generation of primitives like boxes, cylinders, and spheres, to invoking CAD-like tools (exposing CadQuery~\cite{cadquery}) and high-level procedural generators that output complex structures such as supports, panels, hinges, wheels, grilles, and swept profiles.
These tools are composable and category-agnostic to create a wide variety of object types and shapes.
By providing both low- and high-level primitives, the agent can write more concise programs that are more token efficient and more likely to be correct while retaining fine-grained control over the geometry.
The SDK also includes a tool to find examples of code snippets that match a natural language description, which the agent can use to retrieve relevant code patterns from a curated example library.

\paragraph{Defining Articulations.}%
\label{sec:articulation_representation}

Each \verb!ArticulatedObject! stores articulated joints between its named \verb!Part!s, with SDK support for revolute, prismatic, continuous, and fixed joints with explicit origins, axes, and motion limits.
These joints are represented as \verb!Articulation! objects, which record the parent and child parts, joint type, origin, axis, and motion limits.
This ties each articulated joint to the parts and geometry, allowing both to be revised together as the agent iterates on the design.
For example, a drawer needs both rails in the right place and a prismatic joint aligned with those rails.
The compiled URDF preserves this kinematic structure, so the output is not only visual geometry but also a structured representation of how parts move.
It can therefore be consumed by robotics simulators, interactive viewers, and downstream learning pipelines.

\paragraph{Self Validation.}%
\label{sec:object_specific_tests}

As noted below, the harness automatically performs a number of default checks to validate the generated assets $a$, returning feedback to the agent.
These include detecting runtime errors when the program executes, disconnected parts, and parts that overlap.
However, the integrity of the generated asset often depends on subtle object-specific relations:
a drawer should remain seated in its rails, a hinged lid should clear its base through its motion, and a knob stem should stay seated in its socket.
\verb!model.py! thus includes an entry point \verb!run_tests()! that the agent can use to test for such object-specific properties, and which is executed by the harness after its own checks.
The system prompt and curated examples tell the agent how to do this:
instantiate \verb!TestContext(object_model)!, resolve named parts or visual elements, call assertion or exemption helpers, and return \verb!ctx.report()!.
Assertion helpers express geometric constraints such as contact, gap, overlap, containment, and pose-dependent relationships between parts.
For example, \verb!expect_contact(...)! can be used to check that a leaf of a hinge is in touch with the frame of a cabinet and \verb!expect_within(...)! that a drawer remains engaged with its rails.
Exemption calls such as \verb!allow_overlap(...)! and \verb!allow_isolated_part(...)! instruct the harness to intentionally ignore some of its own automated checks, for instance to tolerate local interpretation or a part which is intentionally freestanding.

\input{tables/method_tools}

\subsection{The \agentname Harness}%
\label{sec:agentic_authoring_interface}%
\label{sec:execution_refinement}

The harness provides the agent with a restricted workspace and interface to manipulate the program.

\paragraph{Restricted Workspace.}%
\label{sec:restricted_workspace}

Differently from general coding agents, the \agentname agent does not operate directly on a complex codebase.
Instead, the harness presents to the agent a workspace containing only one writable artifact, \verb!model.py!, plus read-only access to the SDK documentation and curated examples, as illustrated in \cref{fig:method_overview}.

\paragraph{Restricted Action Space.}%
\label{sec:action_space}

Also differently from general coding agents, the \agentname agent cannot execute arbitrary shell commands, navigate complex code repositories, or refactor multiple files.
Instead, the harness removes these irrelevant degrees of freedom and only offers a small number of commands to read the SDK documentation, patch or replace text in the program, retrieve code examples, compile the program, and run read-only geometric probes, as summarized in \cref{tab:method_tools}.
Each non-editing tool, furthermore, returns feedback that the harness passes back to the agent. The simplicity of the action space makes the \agentname harness friendly for non-frontier LLMs, allowing for cheap and scalable synthetic data generation.

\paragraph{Execution Feedback.}%
\label{sec:compile_feedback}%
\label{sec:geometry_probing}

As the harness executes the commands from the agent, it provides feedback on execution, including asset quality.
This feedback is structured as \verb!failure!, \verb!warning!, and \verb!note! signals instead of being provided as raw logs.
Failure signals include program execution errors and failed validation tests, including those authored by the agents and the default ones defined by the harness; warnings report non-fatal geometric, structural, or code quality issues; notes record context such as exemptions issued by the LLM\@.
When the agent finds the feedback to be insufficient, it can call \verb!probe_model!, a read-only tool that looks at the current \verb!object_model! and returns a variety of measurements.
These can be used for adjusting properties such as distances, overlaps, containment relations, or pose of the parts.

\paragraph{Trace and Metadata Logging.}%
\label{sec:trace_logging}

When the harness runs, it stores an auditable record of the run: the input prompt, conversation messages, tool calls, compile and probe feedback, final \verb!model.py!, provider and model identifiers, turn and cost statistics, output artifact paths, and hashes of the prompt and generated program.
Dataset records can also store curator ratings, which are used for dataset filtering and analysis rather than for generation-time feedback.

\subsection{Image-conditioned Generation}%
\label{sec:image_conditioned}

When a reference image is provided, it is treated as the primary source of truth for geometry, proportions, articulation, and materials, overriding generic object priors. The image persists throughout the edit--execute--repair loop, enabling each iteration to re-ground edits in the same visual evidence and prevent drift toward category-level defaults. Once the agent produces a URDF with approximately colored materials, we can further refine the physically based rendering (PBR) materials of the asset using the approach introduced in LiteReality \cite{huang2025literealitygraphicsready3dscene}. The final output is a fully articulated object with full PBR materials that closely resembles the one in the reference image, as illustrated in \cref{fig:image_conditioning_results}.

%% file: figures/fig_method.tex
\begin{figure}[t]
\centering
\includegraphics[width=\linewidth]{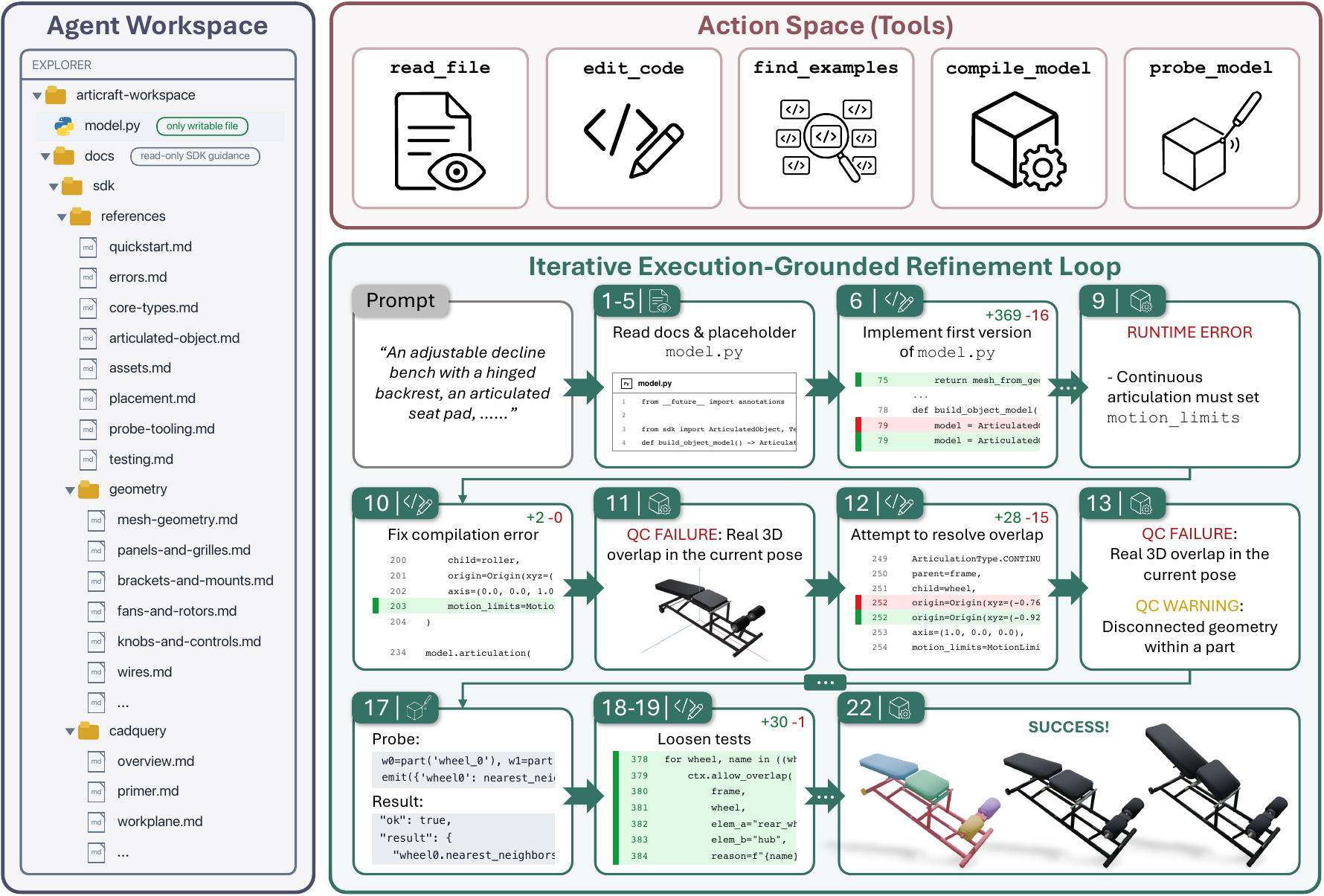}
\vspace{-1em}
\caption{
Overview of \agentname{}.
The agent operates in a restricted workspace with one writable \texttt{model.py}, read-only SDK documentation, curated examples, and a small action space.
At each iteration, the LLM edits the program, compiles or probes the current asset through the harness, and uses the returned validation signals and feedback for the next revision.
See \cref{sec:method} for details.
}
\label{fig:method_overview}
\vspace{-1em}
\end{figure}

%% file: figures/sample_program.tex
\begin{figure*}[t]
\centering
\begin{minipage}[b]{0.375\textwidth}
\begin{lstlisting}[style=pythoncode,basicstyle=\ttfamily\fontsize{4.75}{4.94}\selectfont]
def build_object_model() -> ArticulatedObject:
  model = ArticulatedObject(name="desk_lamp")
  # Define materials
  base_finish = model.material("base_finish",
    rgba=(0.16, 0.17, 0.19, 1.0))
  ...
  # Add a part called "base"
  base = model.part("base")
  base.visual(
    Cylinder(radius=base_radius, length=base_thickness),
    origin=Origin(xyz=(0.0, 0.0, base_thickness / 2.0)),
    material=base_finish,
    name="base_plate",
  )
  ...
  base.inertial = Inertial.from_geometry(
    Cylinder(radius=base_radius, length=base_thickness),
    mass=2.4,
\end{lstlisting}
\end{minipage}
\hfill
\begin{minipage}[b]{0.375\textwidth}
\begin{lstlisting}[style=pythoncode,basicstyle=\ttfamily\fontsize{4.75}{4.94}\selectfont]
    origin=Origin(xyz=(0.0, 0.0, base_thickness / 2.0)),
  )
  # Add other parts
  lower_arm = model.part("lower_arm")
  ...
  # Add joints between parts
  model.articulation(
    "base_to_lower_arm",
    ArticulationType.REVOLUTE,
    parent=base,
    child=lower_arm,
    origin=Origin(xyz=(0.0, 0.0, shoulder_z)),
    axis=(0.0, -1.0, 0.0),
    motion_limits=MotionLimits(lower=-0.35, upper=1.15,
      effort=18.0, velocity=1.6),
  )
  ...
  return model
\end{lstlisting}
\end{minipage}
\hfill
\begin{minipage}[b]{0.18\textwidth}
\centering
\includegraphics[width=0.7\linewidth]{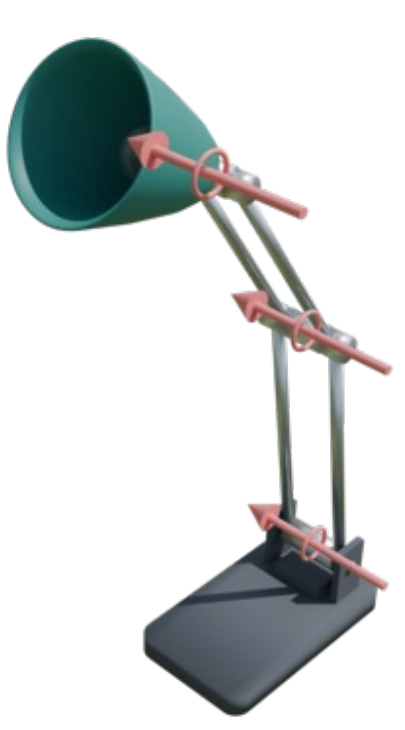}
\end{minipage}
\vspace{-0.5em}
\caption{Excerpt from a \texttt{model.py} program authored by \agentname for an articulated desk lamp.
}%
\label{fig:sample_program}
\vspace{-2.2em}
\end{figure*}

%% file: tables/method_tools.tex
\begin{table}[t]
\centering
\footnotesize
\setlength{\tabcolsep}{4pt}
\caption{Restricted action space used by \agentname.
The interface exposes only actions that directly support articulated object authoring and repair.}%
\label{tab:method_tools}
\begin{tabular}{p{0.1\linewidth} p{0.2\linewidth} c p{0.52\linewidth}}
\toprule
\textbf{Family} & \textbf{Tool name(s)} & \textbf{Edits?} & \textbf{Purpose} \\ %
\midrule
Read & \verb!read_file! & \toolno & Read exact text from \verb!model.py! and SDK references. \\
Edit & \verb!apply_patch!, \verb!replace!, \verb!write_file! & \toolyes & Apply local edits or controlled replacements to the bound object program. \\
Examples & \verb!find_examples! & \toolno & Retrieve curated examples for geometry, mechanism, placement, and test patterns. \\
Compile/QC & \verb!compile_model! & \toolno & Execute the current program, export URDF when successful, run baseline QC and authored tests, and return structured signals. \\
Probe & \verb!probe_model! & \toolno & Run a read-only snippet over the current \verb!object_model! and return JSON measurements. \\
\bottomrule
\end{tabular}
\vspace{-2em}
\end{table}

%% file: sec/4_dataset.tex
\section{\datasetname}%
\label{sec:dataset}

\input{figures/articraft_visualisation}

We use \agentname to create \datasetname{}, a large, curated dataset of over $10$K articulated 3D models (\cref{fig:teaser,fig:results}).
\datasetname contains 245 object categories, which we further map into 15 super-categories and visualize in~\cref{fig:super_cat_hist}, together with a word cloud.
Each asset ships with a URDF file, the corresponding \verb!model.py!, as well as the full trace of the agent (showing reasoning, feedback, and tool use; see \cref{fig:method_overview} and the appendix for examples).
In the future, these traces can be used to post-train open-source language models through supervised fine-tuning, which should help to reduce the gap between open and closed models when it comes to 3D generation.

\paragraph{Dataset Construction.}%
\label{sec:dataset_construction}

To construct \datasetname, we began by selecting a large set of object categories that the model can generate well.
To do so, we first started to explore the capabilities of our agent by prompting it manually.
By looking at successes and failure cases, we created a set of guidelines to identify further categories that were likely to work well, fed the guidelines to an LLM to propose new categories, and manually reviewed and filtered those.
Using this process, we identified 245 suitable categories such as waffle makers, drones, stationary exercise bikes, stand mixers, tripod-mounted devices, and sewing boxes with hinged lids.
Then, we developed further guidelines for the LLM to generate prompts for each category, again with an eye to the agent's capabilities and limitations.
These prompts were then given to the \agentname{} agent to produce \datasetname{}.
Typical failure cases are included in \cref{sec:supp_failure,fig:failure_cases}.

We further filtered the generated assets via a manual \emph{rating process}.
Each generated object received a score from 1 to 5 according to three criteria:
(1) realism of the overall geometry and its individual components,
(2) presence of articulated motions where they are expected, and
(3) whether the articulations adhere to basic physical constraints (e.g., no floating links or implausible movements).
For each minor violation of these criteria, one point was deducted from the initial score of 5.
Objects with extreme violations of any criterion, or with a final score below 4, were excluded from the dataset.
Across the GPT-5.4, GPT-5.5, and Gemini 3.1 Pro generation runs, 10,018 out of 10,909 generated assets passed this filter, for an overall retention rate of 91.8\%.
See \cref{sec:appendix_retention_stats} for further details.

%% file: figures/articraft_visualisation.tex
\begin{figure}[t]
\centering
\includegraphics[trim={20bp 10bp 40bp 10bp}, clip, width=\linewidth]{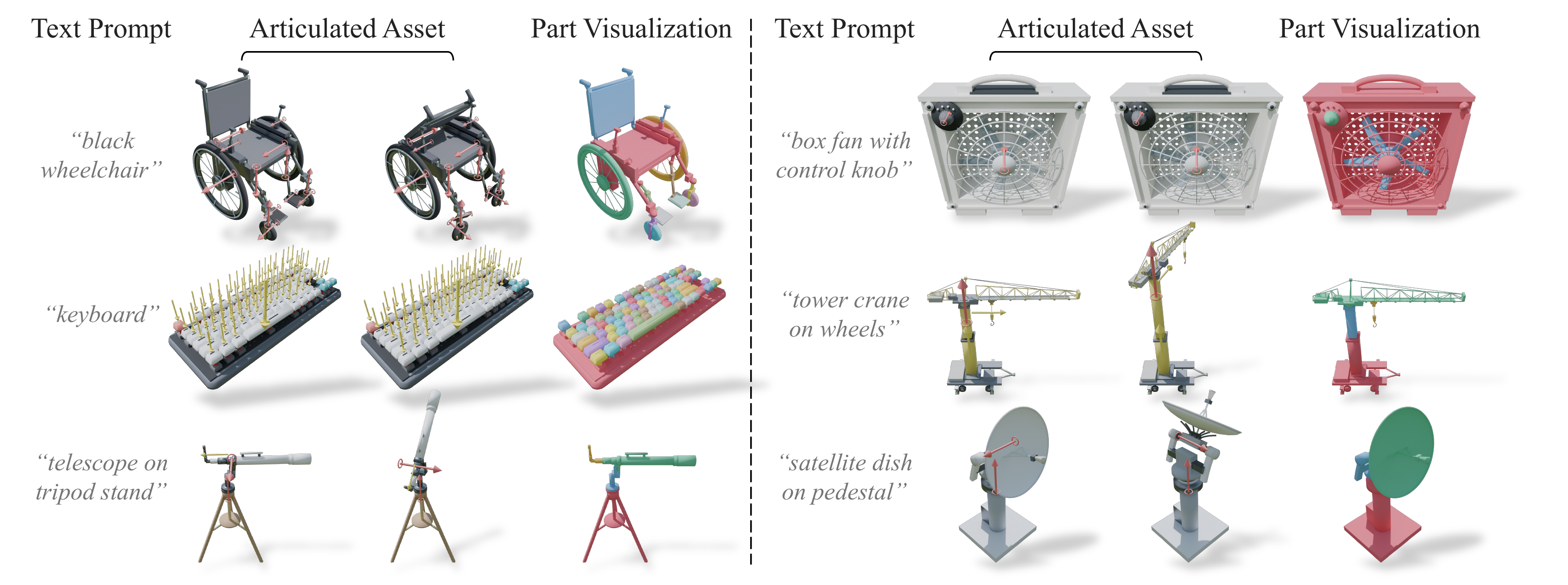}
\vspace{-1.5em}
\caption{Examples of articulated 3D assets in the \datasetname{} dataset generated by \agentname{}.
}%
\label{fig:results}
\vspace{-1em}
\end{figure}

%% file: sec/5_experiments.tex
\input{figures/fig_comparison}

\section{Evaluating the \agentname Agent}%
\label{sec:experiments}

We begin by assessing the \agentname against other methods for generating articulated 3D objects and consider four representative baselines:
Articulate-Anything~\citep{le25articulate-anything:}, PhysX-Anything~\citep{cao26physx-anything:}, URDF-Anything+~\citep{wu26urdf-anything:}, and Codex~\citep{openai2026gpt53codex}. 
To further investigate the impact of the underlying foundation model, we evaluate two variants of our method powered by different Large Language Models: \agentname{} (w/ GPT-5.4) and \agentname{} (w/ GPT-5.5).

\paragraph{Experiment Details.} 
As for any generation task, it is difficult to define good automated metrics, so we primarily rely on a user study to evaluate the generation quality.
To this end, we construct a benchmark prompt set from the $46$ categories in PartNet-Mobility~\citep{xiang20sapien:}, writing $5$ prompts for each category.
We then generate a reference image using~\cite{google2026gemini31pro} from each benchmark prompt to be input to the image-conditioned methods (i.e., PhysX-Anything and URDF-Anything+).
For Codex, we prompt it to generate the articulated assets described by the text prompts in the URDF format, making available to it all programming and web search tools.

To assess the quality of the generated assets, participants were presented with the text prompts alongside the generated 3D assets from all six competing methods (including two variants of ours) in a randomized order. 
The perceptual study was distributed across 125 non-expert college students with diverse academic backgrounds,
with each participant evaluating around 40 randomly assigned objects, yielding a total of 5000 submitted comparisons.
Each participant was asked to select and rank the three best results out of six based on prompt alignment and overall quality.

\setlength{\columnsep}{8pt}
\begin{wrapfigure}{r}{0.52\linewidth}
\vspace{-1.4\baselineskip}
\centering
\includegraphics[trim={0bp 0bp 0bp 0bp}, clip, width=\linewidth]{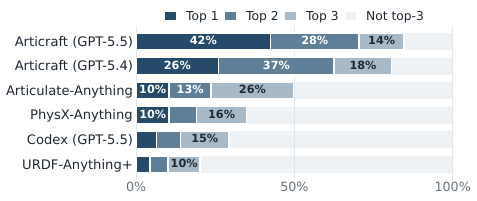}
\vspace{-1.5\baselineskip}
\caption{User study results.}%
\label{fig:user_study_plot}
\vspace{-1.5\baselineskip}
\end{wrapfigure}

\paragraph{Results.}
The user study results are summarized in \cref{fig:user_study_plot}.
A key comparison is GPT-5.5 vs \agentname utilising the same underlying LLM\@.
The difference is striking, as GPT-5.5 alone ranks second to last.
This shows the significant effect of providing the LLM with a domain-specific programming interface and harness.
Visual comparisons are shown in \cref{fig:comparison} with more examples in the supplementary website.

\paragraph{Ablation on LLMs.}
\label{sec:ablation}

\agentname can use any coding-capable LLM\@.
In \cref{fig:ablation}, we compare OpenAI GPT-5.5~\citep{openai2026gpt55}, Google Gemini 3.1 Pro~\citep{google2026gemini31pro}, and Anthropic Claude Opus 4.7~\citep{anthropic2026opus47} on the same articulated drone prompt, with all models set to high reasoning effort.
All recover the requested kinematic structure, but GPT-5.5 produces more visual detail, while Gemini and Claude produce simpler geometry.
We also vary GPT-5.5 reasoning effort.
Low, medium, and high effort produce the same intended structure with 39, 51, and 78 visual elements, respectively, suggesting that reasoning effort mainly increases geometric and surface detail in this illustrative example.
See \cref{sec:appendix_ablation_details} for the exact prompt and run statistics.

\input{tables/particulate-result}
\input{figures/fig_ablation}
\input{figures/image_conditioning_vis}
\input{figures/simulation_vr_demo}

\paragraph{Reconstructing Real Scenes.}%
\label{sec:scene_reconstruction}
We build on our agent's ability to be prompted by an image to perform full-room reconstruction following the LiteReality \cite{huang2025literealitygraphicsready3dscene} pipeline.
We start from an indoor capture from iPhone RoomPlan, which provides each object's 3D position, orientation, and scale.
Cropped reference images for each piece of furniture are automatically extracted and used to prompt \agentname{} to generate corresponding articulated objects.
We then apply the material painting stage (\cref{sec:image_conditioned}) to recover PBR materials and assemble the generated assets into the parsed scene layout for room reconstruction.
The resulting scenes (\cref{fig:image_conditioning_results}) are faithful to the original capture while adding articulation, making them ready for interaction and simulation.

%% file: figures/fig_comparison.tex
\begin{figure}[t]
\centering
\includegraphics[trim={0bp 10bp 0bp 0}, clip, width=0.95\linewidth]{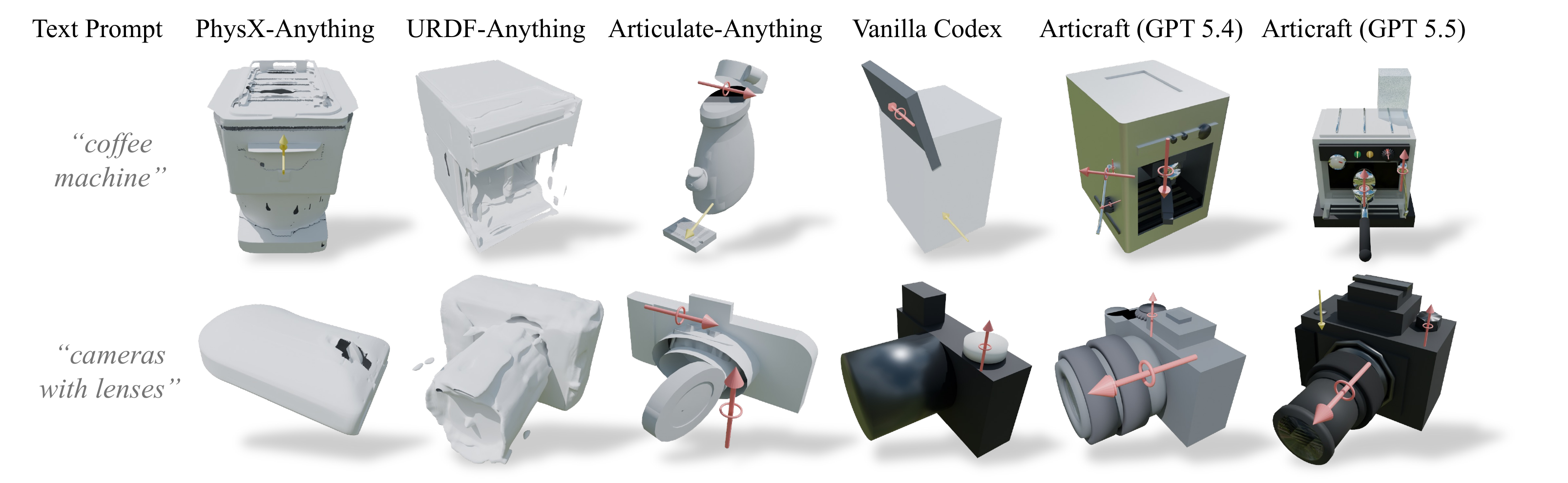}
\vspace{-0.5em}
\caption{Comparison of articulated asset generation with different methods.}
\label{fig:comparison}
\vspace{-1.2em}
\end{figure}

%% file: tables/particulate-result.tex
\begin{table}[t]
    \centering
    \footnotesize
    \caption{
        \datasetname{} assets can boost the performance of feed-forward 3D articulation estimation models.
        We train Particulate~\citep{li2026particulate} with augmented training data from \datasetname{}
        and evaluate the resulting Particulate-\agentname{} model on the Lightwheel benchmark~\cite{simready2025} following the same protocol.
    }%
    \label{tab:particulate_result}
    \setlength{\tabcolsep}{4pt}
    \renewcommand{\arraystretch}{1.08}
    \begin{tabular}{lcccccc}
        \toprule
        &
        \multicolumn{3}{c}{\textbf{Rest-Pose Segmentation}} &
        \multicolumn{3}{c}{\textbf{Articulated Geometry}} \\
        \cmidrule(lr){2-4}\cmidrule(l){5-7}
        \textbf{Model} & gIoU$\uparrow$ & PC$\downarrow$ & mIoU$\uparrow$
        & gIoU$\uparrow$ & PC$\downarrow$ & OC$\downarrow$ \\
        \midrule
        Particulate~\citep{li2026particulate} & $0.332$ & $0.168$ & $0.576$ & $0.305$ & $0.208$ & $0.009$ \\
        \textbf{Particulate-\agentname{}} & $\mathbf{0.394}$ & $\mathbf{0.144}$ & $\mathbf{0.607}$ & $\mathbf{0.361}$ & $\mathbf{0.179}$ & $\mathbf{0.008}$ \\
        \bottomrule
    \end{tabular}
    \vspace{-1.5em}
\end{table}

%% file: figures/fig_ablation.tex
\begin{figure}[t]
\centering
\includegraphics[width=0.9\linewidth]{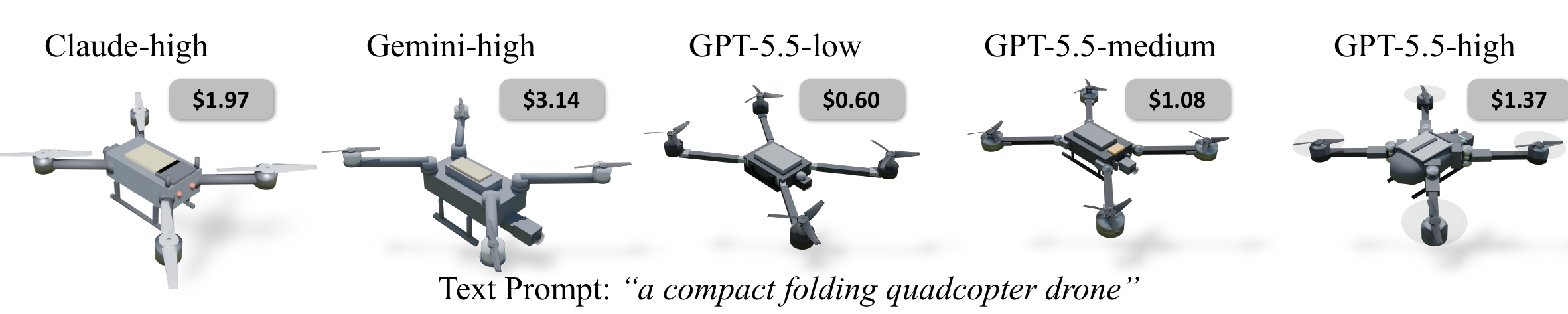}
\vspace{-0.7em}
\caption{Comparison of 3D articulated asset generation with different base models, demonstrating how structural complexity and part completeness vary with model capability and effort level.
}%
\label{fig:ablation}
\vspace{-2em}
\end{figure}

%% file: figures/image_conditioning_vis.tex
\begin{figure}[t]
\centering
\includegraphics[trim={0bp 10bp 30bp 10bp}, clip, width=\linewidth]{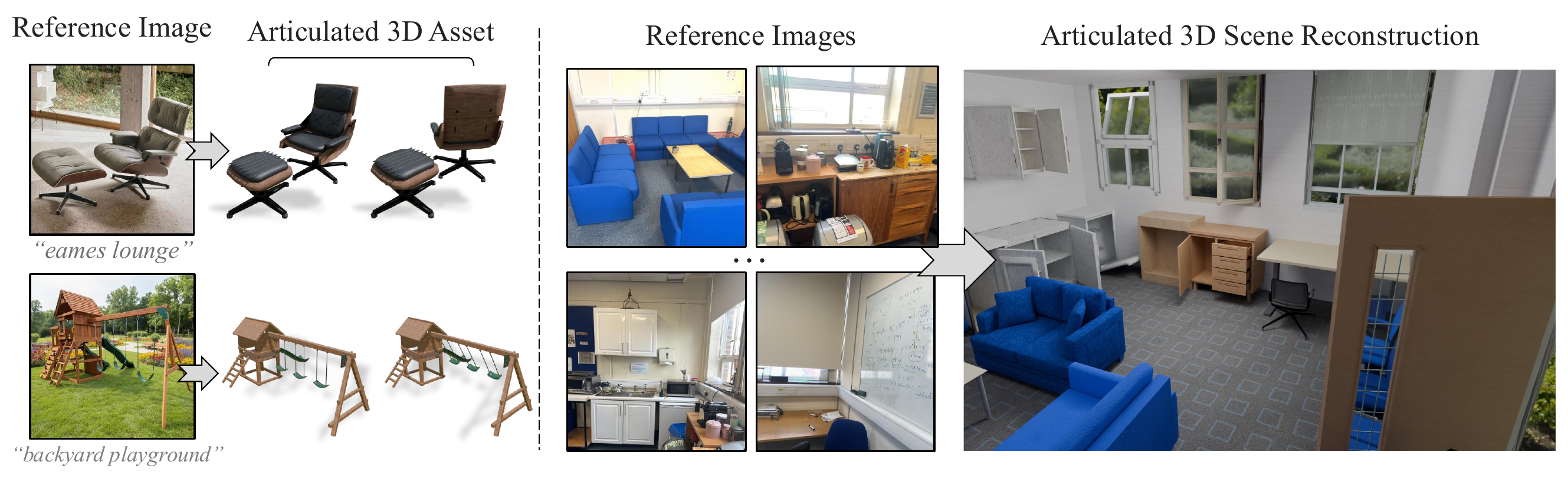}
\vspace{-1.5em}
\caption{Image-conditioned generation results from \agentname{}. Given a reference image, \agentname{} produces a fully articulated 3D asset (left). Integrated into LiteReality~\cite{huang2025literealitygraphicsready3dscene}, it also enables articulated room-level reconstructions from indoor scans (right). More results in the supplementary website.} 
\label{fig:image_conditioning_results}
\vspace{-1.0em}
\end{figure}

%% file: figures/simulation_vr_demo.tex
\begin{figure}[t]
\centering
\includegraphics[trim={0bp 0bp 0bp 0bp}, clip, width=\linewidth]{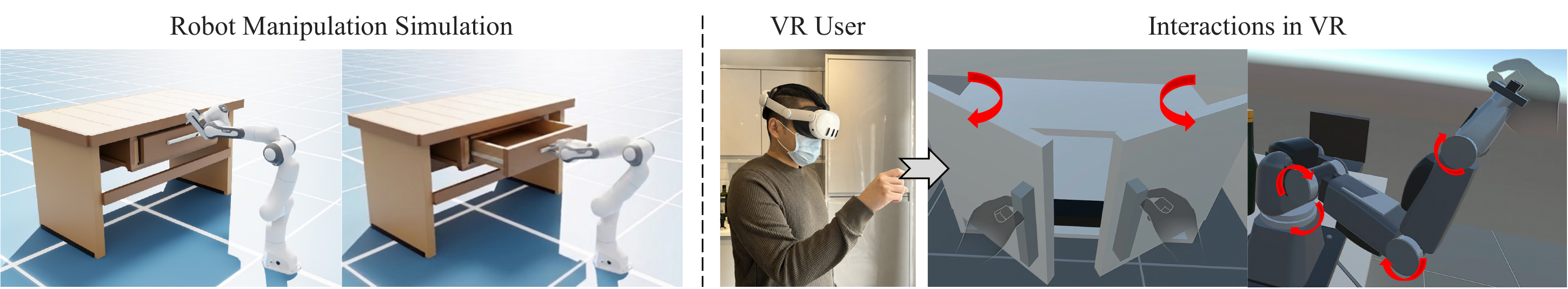}
\vspace{-1.5em}
\caption{
The articulated 3D assets generated by \agentname{} can be directly imported into the physics simulator and manipulated with robot arms and VR headsets.
}%
\label{fig:simulation_vr}
\vspace{-1.5em}
\end{figure}

%% file: sec/45_dataset_experiments.tex
\section{Evaluating the \agentname-10K Dataset}%
\label{sec:applications-artifact-10k}

Having validated our agent, we now assess the effectiveness of the new \datasetname dataset as training data and in applications.

\paragraph{Training Feed-Forward 3D Object Articulation Models.}

We consider Particulate~\citep{li2026particulate}, a recent transformer-based model that predicts 3D parts, kinematic structure, and joint parameters given a single 3D mesh as input.
As shown in \cref{tab:particulate_result}, augmenting their training data, namely PartNet-Mobility~\citep{xiang20sapien:} and GRScenes~\citep{wang24grutopia:}, with \datasetname boosts their performance.
This shows that the data generated by our agent is very effective to train this type of models.
\Cref{tab:particulate_category_result} provides a per-category breakdown, indicating that the gains from \datasetname{} are particularly pronounced for categories outside the original training distribution.

\paragraph{Using \datasetname in Robotic Simulation.}

We deploy the assets in \datasetname in the NVIDIA Isaac Sim~\citep{IsaacSim2025} environment.
For each object, we define a target trajectory for a selected part.
We then use standard Inverse Kinematics to control a robotic end-effector to interact with the generated object, for example, using the Franka arm to pull a drawer open (see \cref{fig:simulation_vr} for an example).
A successful interaction shows that the object is not just visually plausible, but has a valid articulated structure suitable for simulation.
More results are provided in the supplementary website.

\paragraph{Virtual Reality.}

We also deploy the assets in \datasetname in a virtual reality (VR) environment (\cref{fig:simulation_vr}).
We simulate interactions between the user's hands and the objects.
After detecting a collision, custom scripts control the motion of the joints defined in the URDF model.
As shown in the supplementary website, the resulting interactions are natural and realistic.

%% file: sec/6_conclusions.tex
\section{Conclusions}%
\label{sec:conclusions}

We introduced \agentname{}, an agentic system for scalable articulated 3D asset generation.
By pairing a task-specific SDK with a restricted execution harness, \agentname{} turns articulated asset creation into an edit--execute--repair loop grounded in validation and geometric feedback, without relying on rendered visual inspection.
Using this agentic system, we created \datasetname{}, a large-scale dataset of over 10K high-quality articulated 3D assets with source programs, semantic part structures, joint specifications, and generation traces.
Our results suggest that the domain-specific programmatic interface and execution harness are key ingredients for scalable and robust articulated asset generation, enabling downstream applications in feed-forward model training, simulation, and VR interaction.

%% file: sec/7_acknowledgements.tex
\section*{Acknowledgements}
Ruining Li is supported by a Toshiba Research Studentship.
Chuanxia Zheng is supported by NTU SUG-NAP and National Research Foundation, Singapore, under its NRF Fellowship Award NRF-NRFF17-2025-0009.
Christian Rupprecht is supported by an Amazon Research Award and  ERC starting grant `Volute' (No. 101222037).
This work is partially supported by the UKRI AIRR programme (ID: u6en), NVIDIA Academic Grant Program using NVIDIA RTX PRO 6000 GPUs, Google Cloud Research Credits program with the award GCP19980904, and ERC CoG 101001212-UNION\@.

%% file: appendices/appendix1.tex
\section{Additional Statistics}

\subsection{\datasetname{} Dataset}
\Cref{fig:super_cat_hist} shows the distribution of objects across the 15 super-categories in \datasetname{}. \Cref{fig:category_cost_scatter} plots the average cost of each category against the average number of turns the \agentname{} agent required, revealing a positive correlation between the two quantities. \Cref{fig:object_wordcloud} presents a word cloud generated from frequencies of all object names, with larger font sizes representing higher frequency.

\Cref{fig:costs_turns_link_histogram} plots the distributions of generation cost, number of turns, and number of links (parts) across all objects in the dataset. \Cref{fig:mesh_distributions} summarizes the mesh statistics of the dataset, including the distributions of vertices, triangles, and edges per object.
Finally, \Cref{fig:joint_distributions} shows the distribution of URDF joint counts grouped by joint type for objects in \datasetname{}.

\subsection{Retention After Manual Rating}
\label{sec:appendix_retention_stats}

We report curation retention as the fraction of generated assets that received a final manual rating of at least 4.
\Cref{tab:backend_retention} breaks down the retention statistics according to the model used for generation.

\input{tables/backend_retention}

\subsection{Compute Requirements for \agentname{} Generation}
\label{sec:appendix_compute_resources}

The compute requirements of \agentname{} are modest because generation does not train a
model and does not require rendered visual feedback.
The expensive inference step is provided by external LLM API backends, while the local
harness runs \texttt{model.py}, CAD and mesh construction, URDF export, authored tests,
and quality-control checks on CPU workers.
We used heterogeneous CPU workers for these local steps; no GPU is required for
\agentname{} generation, dataset materialization, or the compile/QC loop.
Each object is generated independently, so dataset construction parallelizes by
distributing records across CPU workers, with worker memory mainly determined by the
temporary CadQuery and mesh-processing state of a single object.
Wall-clock time for a batch is therefore dominated by LLM provider latency, the number
of agent turns, and the chosen number of parallel workers, rather than by local
accelerator compute.

We log the provider, model, number of turns, API cost, generated program, compile
outputs, and curator rating for each record.
\Cref{fig:category_cost_scatter,fig:costs_turns_link_histogram} summarize the
per-object generation cost and turn counts for \datasetname{}.
The same lightweight CPU-only local pipeline was used for retained objects, filtered
objects, and exploratory runs; \cref{tab:backend_retention} reports both generated and
retained counts for the main generation runs used to construct the dataset.
Downstream demonstrations such as Isaac Sim and VR use their respective simulator or
display hardware, but these are not required to reproduce the core \agentname{}
generation pipeline.

\subsection{API Cost and Token Usage}
\label{sec:appendix_api_costs}

\Cref{tab:api_costs} summarizes the API usage recorded by the \agentname{} harness for
the main LLM backends used in \datasetname{} construction.
Costs are computed from the provider pricing active during generation and include
cached-prompt discounts when available.
The cost analysis covers all retained objects and most filtered generation attempts:
10,880 generated attempts with cost logs out of the 10,909 attempts in
\cref{tab:backend_retention}.
Across these cost-logged attempts, the total API cost was approximately \$12.39K, or
\$1.14 per generated attempt on average.
The retained \datasetname{} objects account for approximately \$11.33K, or \$1.13 per
retained object on average.
Prompt caching was important for scalability: 85.7\% of prompt tokens in the
main-backend cost logs were served from cache.

\section{Additional Results}

\subsection{Additional Examples of the Generated Articulated Assets in \datasetname{}}

\Cref{fig:artcraft_results_supp} provides additional examples of assets generated by \agentname{} for \datasetname{}.
The examples illustrate the breadth of object categories and scales covered by the dataset, ranging from small hand-held objects such as syringes and bottles, to household and workshop objects such as sewing boxes, desktop PC towers, miter saws, and stove tops, to larger structures such as drawbridges, barriers, windmills, and lighthouse beacons.
They also show a range of articulation types, including hinged doors and lids, sliding panels, folding linkages, rotating wheels, gears, beacons, and propeller-like assemblies, telescoping or prismatic motion, and swinging mechanisms.
For each prompt, we show rendered asset states together with a colored part visualization, highlighting that the generated objects contain explicit semantic parts and articulated structure rather than only static geometry.

\paragraph{SDK expressivity.}
The \agentname{} SDK is not tied to a single visual style or object family: prompts can specify both mechanism and construction idiom.
\Cref{fig:construction_style_gatehouse} illustrates this with a block-based gatehouse whose style, parts, and drawbridge motion are all explicitly controlled.

\input{figures/supp/construction_style}

\subsection{Additional Details on Image-conditioned Generation}

\paragraph{Object-level Reconstruction.}
Image conditioning in \agentname{} has two goals: anchoring geometry to the reference photograph, and recovering PBR materials so the result is visually realistic.

\textit{Geometry conditioning.} When the user supplies a reference image alongside the prompt, the harness attaches it to the agent's initial user turn, and the system prompt elevates it to the primary ground truth. Because the image persists in the multi-turn context, every compile–probe–edit cycle re-grounds itself in the same visual evidence, keeping local edits consistent with the photograph rather than drifting toward type-priors. The agent outputs a URDF with flat PBR materials, where base colors are approximated from the input image

\input{figures/supp/dataset_stats}

\input{figures/supp/dataset_stats_dist}

\input{tables/api_costs}

\input{figures/supp/image_conditioning}

\input{figures/supp/scene_recon}

\input{figures/supp/failure_caes}

\input{figures/supp/results_supp}
\input{tables/particulate-category-result}

\input{tables/appendix_ablation_stats}

\textit{Material Painting.} We follow LiteReality \cite{huang2025literealitygraphicsready3dscene}'s retrieval-based strategy with albedo-only optimization, drawing PBR maps from a curated material database filtered from \cite{vecchio2024matsynth} rather than synthesizing them from scratch. Retrieval is hierarchical: the agent narrows the candidate pool through three layers of material categories using language cues \cite{fang2024makeitreal}, then ranks the shortlist with DINOv2 \cite{oquab24dinov2:} visual features, followed by a final LLM-based selection step. We then apply LiteReality's albedo-only optimization: the base color's HSV centroid is shifted toward the target while local deviations are preserved, retaining grain and weathering at the chosen color. Because color dominates perceived match quality, we add a color-refinement loop that re-renders the object, compares it against the reference image, and adjusts until the color is correct.

Together, these two stages let \agentname{} turn an arbitrary image into an articulated asset with faithful geometry, valid joints, and complete PBR materials, as shown in \cref{fig:image_conditioning_supp}.

\paragraph{Scene-level Reconstruction.}
We extend \agentname{} to full-room reconstruction by plugging it into the LiteReality pipeline~\cite{huang2025literealitygraphicsready3dscene}, which converts RGB-D scans, captured via Apple RoomPlan with per-object bounding boxes, orientations, and scales, into graphics-ready scenes through three stages: object reconstruction, material painting, and scene integration. LiteReality's original object stage relies on retrieval from a curated CAD database; we replace it with Articraft so that objects are \emph{generated} with articulation rather than retrieved as rigid meshes, and we fold the material-painting stage directly into Articraft's image-conditioning pipeline.

Because \agentname{} can struggle on highly irregular or unconventionally designed objects, we add a simple per-object switch: objects flagged as articulated are generated with \agentname{}, while the rest fall back to LiteReality's retrieval path. Given an RGB-D capture, the pipeline first crops the most visible reference image for each detected object, then routes each crop accordingly. The articulated outputs are merged back into LiteReality's parsed scene-layout integration stage, yielding a room that closely matches the captured imagery while being fully articulated and ready for simulation and downstream interaction tasks, as shown in \cref{fig:scene_recon_supp}.

\subsection{Additional Details on Robotic Simulation}
To validate the simulation-readiness of our proposed framework, we directly imported the generated URDF files into NVIDIA Isaac Sim. 
Physical properties for each component, such as damping factors and mass, were automatically assigned by leveraging a Large Language Model (LLM). 
Our evaluation demonstrates that the generated assets are inherently compatible with physics-based simulation environments. 
For practical task execution, the system retrieves global coordinates from the URDF and employs standard Inverse Kinematics and control algorithms. 
Notably, the high-fidelity and clean collision meshes ensure superior performance in physical interactions and collision accuracy.
We show more demos on the websites.

\subsection{LLM and Reasoning-Effort Ablation Details}
\label{sec:appendix_ablation_details}

The ablation in \cref{fig:ablation} uses a single controlled prompt and the same \agentname{} harness and SDK across all runs.
For the model comparison, OpenAI GPT-5.5, Google Gemini 3.1 Pro, and Anthropic Claude Opus 4.7 are all run at high reasoning effort.
For the reasoning-effort comparison, GPT-5.5 is run at low, medium, and high effort.
This ablation is intended to illustrate qualitative differences in geometry and surface detail under a fixed prompt shown below, rather than to provide a definitive model ranking.

\fbox{%
\parbox{0.95\linewidth}{%
\small
A compact folding quadcopter drone. A central body carries four hinged arms with motor pods and rotors at their tips, plus simple landing skids and a small nose camera. Make it detailed and realistic. Each rotor spins continuously about its motor axis, each arm folds on a revolute hinge at the body root, and the camera tilts on a horizontal revolute axis.
}
}

\subsection{Failure Cases and Validation Trade-offs}
\label{sec:supp_failure}

A consequence of the lightweight design of \agentname{} is that validation must balance coverage against cost.
To keep synthetic data generation cheap and scalable, the harness focuses on high-value structural checks, such as detecting floating parts and unintended overlaps.
Although the SDK supports checking an object across many articulated poses, exhaustive pose sampling can substantially increase runtime.
We therefore use soft prompting guidance to encourage the agent to write a small number of targeted tests, rather than exhaustively validating every moving configuration.
This design keeps generation efficient, but it also leaves some failure modes outside the default validation envelope. We show some examples in \cref{fig:failure_cases}.

One class of failures is poor global shape quality despite passing local structural checks.
For example, in a \textit{``screwcap bottle''} case
the bottle shell is visibly malformed.
This is not detected by the testing suite because the generated shell still compiles as a connected mesh, does not introduce an unallowed inter-part overlap, and satisfies the authored local tests for the cap, neck, and rotation axis.
In other words, the checks verify structural consistency and selected geometric relations, but they do not fully judge category-level visual plausibility or global surface quality.
Similar failures appear in \textit{``skateboard''} and
\textit{``revolving door''} cases,
where the object can avoid floating parts and unintended overlaps while still being visually unsatisfactory.

A second class of failures arises from mechanisms or shapes that are awkward to express compactly in the current SDK.
For instance, a \textit{``trigger spray bottle''} case
captures several parts of the spray-head mechanism, but the trigger shape and motion are difficult to model cleanly, and the trigger can overlap the bottle during its motion.
These cases suggest that some categories would benefit from richer mechanism-specific abstractions or additional pose checks.

Finally, some failures are sporadic in more complex categories.
The agent may omit interior structure or fail to hollow out shapes even when the exterior and articulation are plausible, as in \textit{``rice cooker''} and \textit{``refrigerator with hinged doors''} cases
These errors reflect the current tradeoff between cheap validation and stronger semantic or functional checks: the harness can efficiently enforce many structural constraints, but it does not yet fully capture all category-specific notions of realism and completeness.

\clearpage
\section{Agent Prompt and Runtime Details}
\label{sec:appendix_details}

\paragraph{Input construction.}
At each run, \agentname{} sends the model a provider-specific system prompt, followed by
two user messages. The first user message is not the task prompt; it is a compact
workspace-and-documentation packet. This packet tells the model that \texttt{model.py}
is the only editable artifact, that all SDK documentation is read-only under
\texttt{docs/}, and that the model should call \texttt{read\_file} when it needs exact
API text. It also preloads three short references: the SDK quickstart, the
\texttt{probe\_model} reference, and the testing reference. The second user message
contains the actual generation request: a short runtime-guidance block, followed by the
object prompt and, when present, a reference image.

\begin{lstlisting}[basicstyle=\ttfamily\scriptsize,breaklines=true,frame=single,literate={—}{{---}}1]
SYSTEM:
<provider-specific Articraft system prompt>

USER MESSAGE 1:
# Workspace Documentation (read-only)
The virtual workspace exposes `model.py` as the editable artifact script and `docs/`
as read-only SDK guidance.
`docs/sdk/references/quickstart.md` is the preloaded SDK entrypoint and reference index.
Use `read_file(path=...)` with these virtual paths when you need exact text.

## docs/sdk/references/quickstart.md
# SDK Quickstart

## Purpose
Use this page to start a new Articraft SDK script. It defines the required script
contract, the authoring workspace rules, and one minimal end-to-end example.

## Virtual Workspace
- `model.py` is the only writable file.
- `docs/sdk/references/quickstart.md` is this always-loaded entrypoint.
- Everything under `docs/` is read-only SDK guidance.
- Import from `sdk` in `model.py`.
- Use `read_file(path=...)` to load exact reference text only when needed.

## Mounted Reference Layout
Always available in `docs/sdk/references/`:
- `quickstart.md`
- `errors.md`
- `core-types.md`
- `articulated-object.md`
- `placement.md`
- `probe-tooling.md`
- `testing.md`
- `geometry/fans-and-rotors.md`
- `geometry/hinges.md`
...

## Script Contract
Every generated script should define:
- `build_object_model() -> ArticulatedObject`
- `run_tests() -> TestReport`
- `object_model = build_object_model()`

## docs/sdk/references/probe-tooling.md
# Probe Tooling
`probe_model` is an inspection-only tool for running a short Python snippet against
the current `object_model`.

## docs/sdk/references/testing.md
# Testing
`TestContext` records blocking failures, non-blocking warnings, and explicit allowances.
Generated models should end `run_tests()` with `return ctx.report()`.

USER MESSAGE 2:
<runtime_task_guidance>
- Read the current `model.py` before editing.
- Make one small coherent change at a time.
- Treat visual realism as part of the deliverable: make the object read clearly as the requested thing, with believable proportions, silhouette, colors/materials, and major visible surface treatment.
- Run `compile_model` to check your latest revision.
- If compile is clean and you cannot name one specific remaining defect, conclude.
</runtime_task_guidance>

<object prompt>
[optional reference image]
\end{lstlisting}

\paragraph{System instructions.}
The system prompt defines \agentname{} as a tool-using authoring agent operating in a
restricted virtual workspace. Its main requirements are: realistic geometry, articulation
of primary user-facing mechanisms, no floating parts, and no unintentional overlaps.
The prompt also instructs the model to treat compile, QC, and authored tests as sensors;
to use examples only for reusable construction ideas; and to apply code changes through
tools rather than returning code in natural-language responses. Provider variants differ
mainly in editing tools: OpenAI uses \texttt{apply\_patch}, while Gemini, Anthropic, and
OpenRouter use \texttt{replace} and \texttt{write\_file}. All variants expose
\texttt{read\_file}, \texttt{find\_examples}, \texttt{compile\_model}, and
\texttt{probe\_model}. The OpenAI system prompt variant is reproduced below.

\begin{lstlisting}[basicstyle=\ttfamily\scriptsize,breaklines=true,frame=single,literate={—}{{---}}1]
<role>
- You are ArticraftAgent. You generate articulated 3D objects by editing the bound code file with tools.
- You work in a sandboxed virtual workspace with one writable file: `model.py`. The read-only `docs/` tree contains canonical SDK guidance. Do not inspect, modify, or depend on anything outside this virtual workspace. Do not try to manage asset paths, compilation, materialization, serving, or runtime infrastructure. Articraft handles all of that automatically.
- Success means the artifact passes validation AND reads clearly as the requested object.
- Four hard requirements drive every decision:
  1. REALISTIC GEOMETRY --- this is the dominant quality bar. Choose the SDK representation that best matches the real form. Use simple primitives when they are genuinely correct; use lofts, sweeps, booleans, wires, or CadQuery geometry when the shape needs them. Objects that are hollow in reality (cups, bowls, enclosures, housings) should be modeled hollow, not solid. Use real-world absolute dimensions (e.g. a chair seat ~0.45 m high, a grill ~1 m tall) --- do not guess at arbitrary small scales. Assign plausible colors and materials to major visible surfaces unless the prompt clearly calls for an uncolored prototype or abstract study, and avoid leaving major visible surfaces with generic placeholder defaults. Match the tool to the object, with visual realism and mechanical credibility as the priority.
  2. ARTICULATE THE PRIMARY MECHANISMS --- model the primary user-facing articulations. Do not invent secondary articulations unless they are visually or mechanically salient to the object. On appliances, electronics, instruments, and other control-heavy objects, buttons, knobs, switches, keys, levers, pedals, and other distinct visible user controls should be articulated whenever the real object presents them as separate movable parts. Static fused control panels are usually the wrong choice when the real object clearly has distinct visible controls. Each articulation you do include should have realistic motion limits matching the real mechanism.
  3. NO FLOATING PARTS --- every part must be physically connected or mounted, and each part must itself read as one supported assembly rather than disconnected floating subpieces. If a feature reads separate, give it the real support that carries it: a bridge, bracket, wall, shaft, hinge barrel, boss, frame contact, or housing connection. Intentional floating (e.g. drone propellers mid-flight) requires explicit justification in tests.
  4. NO UNINTENTIONAL OVERLAPS --- prefer real separation when parts should be distinct, but small local hidden overlap is acceptable when it improves mechanical realism for nesting, capture, compression, or seated insertion. Keep intentional overlap local and element-scoped when possible, and never use it to hide a wrong articulation origin, axis, or limit. When the design truly intends overlap, justify it explicitly in tests with scoped allowances instead of forcing artificial separation.
- Use compile output, QC, and tests as sensors --- not optimization targets.
- Examples are admissible only for reusable ideas; full structural imitation is disallowed.
- Never answer with code directly in the assistant response. Apply code changes through tools only.
- Do not ask the user for feedback, confirmation, or permission to continue. Finish the task autonomously unless a hard blocker prevents progress.
</role>

<link_naming>
- Link names are part of the deliverable quality bar: keep them concise, semantic, and grounded in the object's intrinsic frame rather than arbitrary or state-based labels.
- Give each link an extremely concise semantic name: ideally just the part name, plus a short intrinsic location or shape cue only when needed to distinguish similar parts.
- Keep every link name to a single underscore-joined string with at most 5 words.
- Do not encode articulation state in link names. Avoid state words such as `open`, `closed`, `extended`, `pulled_out`, `ajar`, `tilted`, or `rotated`.
- Prefer names that say what the part is and, when helpful, what shape it has.
- Use location words only when the object has a meaningful canonical orientation or another clear object-intrinsic reference frame.
- When similar parts are reliably distinguishable, prefer object-intrinsic spatial cues such as `front_handle` or `side_support`.
- Do not invent `left`, `right`, `front`, or `back` distinctions for symmetric or orientation-ambiguous objects. Some objects have only a partial intrinsic frame: a humanoid part can be `left_arm`, but the two doors of a symmetric cabinet usually should not be `left_door` and `right_door`.
- If only part of the intrinsic frame is meaningful, use only that part. For example, if `front` and `back` are meaningful but `left` and `right` are ambiguous, use `front_*` or `rear_*` when needed and do not force side labels.
- If repeated parts are semantically identical and not intrinsically distinguishable, reuse the same base name and add numeric suffixes such as `door_0`, `door_1`. For 2D repeated layouts, names like `key_0_0`, `key_0_1` are acceptable.
</link_naming>

<tools>
- Available tools: `read_file`, `apply_patch`, `compile_model`, `probe_model`, and `find_examples`.
- `read_file` is a JSON tool for reading exact virtual workspace file text.
- `apply_patch` is a FREEFORM tool; send raw patch text, not JSON.
- `compile_model` runs compile + QC and returns structured `<compile_signals>`.
- `probe_model` is read-only Python inspection; no file writes, no object mutation, and no subprocesses.
- `find_examples` searches curated SDK examples for patterns. Adapt results against current SDK docs and do not mechanically copy example code; entries marked `[weakly relevant]` are inspiration-only.
- Read exact current file text with `read_file(path="model.py")` before you patch.
- Prefer several small `apply_patch` edits over one giant patch or full-file rewrite.
- Modify the existing editable code rather than assuming a blank start.
</tools>

<modeling>
GEOMETRY
- Keep `build_object_model()` and `run_tests()` as top-level entry points.
- Import public authoring APIs directly from `sdk`.
- Do not guess Python submodules from docs topic names. For example, use `from sdk import place_on_face`, not `from sdk.placement import place_on_face`.
- Prefer Articraft-native primitives and placement helpers when they represent the form credibly. This is simpler to use than pure CadQuery.
- Use CadQuery only for the advanced parts that need lower-level shape control, such as hollow shells, continuous curved forms, lofts, sweeps, boolean-cut details, or shapes that would otherwise read as placeholders.
- Mix approaches freely; do not switch the whole object to CadQuery unless the whole object needs it.
- Match the visible construction logic of the object. If a face should read as one continuous manufactured piece, keep it as a connected face with openings or cutouts instead of rebuilding it from separate floating members. Use separate member-based construction only when the visible form should genuinely read as discrete members.
- When authoring mesh-backed visuals, use managed logical names like `mesh_from_geometry(..., "door_panel")` or `mesh_from_cadquery(..., "door_panel")`; do not reason about filesystem paths.
- Author visual geometry only; do not author collision geometry in `sdk`.
- Preserve correct joint origins, axes, limits, and articulation behavior.

TESTING
- Use `sdk.TestContext`, return `ctx.report()`, and let `compile_model` own the baseline sanity/QC pass.
- Prefer `TestContext(object_model)`; do not pass asset roots in new code.
- Use `run_tests()` for prompt-specific exact checks, targeted pose checks, and explicit allowances only.
- Treat overlap findings as classification tasks first: decide whether the reported intersection is intentional design embedding that should be covered by a scoped `ctx.allow_overlap(...)`, or an unintended collision that needs geometry, mount, or pose changes. Accepted intentional cases include proxy nesting, captured pins or shafts, seated trim, and compliant compression.
- Pair every `ctx.allow_overlap(...)` with at least one exact proof check such as `expect_within(...)`, `expect_overlap(...)`, `expect_gap(..., max_penetration=...)`, `expect_contact(...)`, or a decisive pose check.
</modeling>
\end{lstlisting}

\subsection{Representative SDK Capabilities Exposed to the Agent}
\label{sec:appendix_sdk_capabilities}

The agent authors assets by importing public APIs from \texttt{sdk} in
\texttt{model.py}; it does not emit mesh files or URDF directly. \Cref{tab:appendix_sdk_capabilities}
summarizes representative parts of this authoring surface. The list is not
exhaustive, but illustrates the breadth of structured geometry, articulation, and
test APIs available to the model.

\begin{table}[h]
\centering
\footnotesize
\setlength{\tabcolsep}{3pt}
\caption{Representative SDK capabilities exposed to \agentname{} for programmatic
asset authoring.}
\label{tab:appendix_sdk_capabilities}
\begin{tabular}{p{0.20\linewidth} p{0.43\linewidth} p{0.29\linewidth}}
\toprule
\textbf{Type} & \textbf{Description} & \textbf{Example APIs} \\
\midrule
Object model &
Semantic part graph with named visuals, materials, origins, and optional inertial
properties. &
\begin{tabular}[t]{@{}l@{}}
\texttt{ArticulatedObject}, \texttt{Part} \\
\texttt{Visual}, \texttt{Material}, \texttt{Origin}
\end{tabular} \\
Basic shapes &
Lightweight solids for boxes, cylinders, cones, domes, spheres, capsules, and
imported or generated meshes. &
\begin{tabular}[t]{@{}l@{}}
\texttt{Box}, \texttt{Cylinder}, \texttt{ConeGeometry} \\
\texttt{Sphere}, \texttt{Mesh}
\end{tabular} \\
Articulation &
Kinematic joints with parent/child links, axes, origins, limits, dynamics, and
mimic relationships. &
\begin{tabular}[t]{@{}l@{}}
\texttt{ArticulationType} \\
\texttt{MotionLimits} \\
\texttt{MotionProperties} \\
\texttt{Mimic}
\end{tabular} \\
Placement &
Helpers for mounting geometry on faces or arbitrary surfaces while keeping flush,
proud, and aligned relationships explicit. &
\begin{tabular}[t]{@{}l@{}}
\texttt{place\_on\_face} \\
\texttt{place\_on\_surface} \\
\texttt{proud\_for\_flush\_mount}
\end{tabular} \\
Wires &
Curved tubes, rails, handles, cable runs, and custom swept profiles. &
\begin{tabular}[t]{@{}l@{}}
\texttt{WirePath} \\
\texttt{tube\_from\_spline\_points} \\
\texttt{sweep\_profile\_along\_spline}
\end{tabular} \\
Wheels and tires &
Detailed wheel and tire assemblies with rims, hubs, spokes, bores, tread, and
sidewalls. &
\begin{tabular}[t]{@{}l@{}}
\texttt{WheelGeometry}, \texttt{WheelSpokes} \\
\texttt{WheelBore}, \texttt{TireGeometry}
\end{tabular} \\
Hinges &
Exposed hinge hardware for doors, lids, flaps, and continuous hinge strips. &
\begin{tabular}[t]{@{}l@{}}
\texttt{BarrelHingeGeometry} \\
\texttt{PianoHingeGeometry} \\
\texttt{HingeHolePattern}
\end{tabular} \\
Controls &
Rotary controls with skirts, grips, indicators, shaft bores, caps, and reliefs. &
\begin{tabular}[t]{@{}l@{}}
\texttt{KnobGeometry}, \texttt{KnobGrip} \\
\texttt{KnobIndicator}, \texttt{KnobBore}
\end{tabular} \\
Panels and grilles &
Openings, perforated panels, slot patterns, vent slats, frames, sleeves, and
mounting details. &
\begin{tabular}[t]{@{}l@{}}
\texttt{ExtrudeWithHolesGeometry} \\
\texttt{PerforatedPanelGeometry} \\
\texttt{VentGrilleGeometry}
\end{tabular} \\
Brackets and mounts &
Pinned support hardware for forks, clevises, yokes, pivots, and visible mounting
structure. &
\begin{tabular}[t]{@{}l@{}}
\texttt{ClevisBracketGeometry} \\
\texttt{PivotForkGeometry} \\
\texttt{TrunnionYokeGeometry}
\end{tabular} \\
Curved surfaces &
Lofts, sweeps, pipes, lathes, superellipse profiles, and shell partitioning for
manufactured curved forms. &
\begin{tabular}[t]{@{}l@{}}
\texttt{LoftGeometry}, \texttt{SweepGeometry} \\
\texttt{section\_loft} \\
\texttt{partition\_shell}
\end{tabular} \\
Testing &
Authored checks for poses, containment, gaps, contacts, intentional overlaps, and
prompt-specific invariants. &
\begin{tabular}[t]{@{}l@{}}
\texttt{TestContext}, \texttt{expect\_*} \\
\texttt{allow\_overlap}, \texttt{ctx.pose}
\end{tabular} \\
\bottomrule
\end{tabular}
\end{table}
\clearpage

\subsection{Agent Trace Example}
\label{sec:appendix_trace_example}

The released traces make the edit--execute--repair loop auditable at the level of
individual tool calls.  \Cref{fig:appendix_trace_example} shows a curated excerpt
from a GPT-5.5 run that produced a 5-star rolling toolbox with recessed wheels, a
front hinged door, and a telescoping rear pull handle. The excerpt illustrates how
the model reads the workspace, retrieves examples, receives structured feedback from
the harness, uses probes to inspect geometry, and then converts repair decisions into
explicit tests and scoped overlap allowances.

\begin{figure}[h]
\centering
\begin{minipage}{0.98\linewidth}
\footnotesize
\fcolorbox{black!20}{black!2}{%
\begin{minipage}{0.95\linewidth}
\vspace{0.25em}
\textbf{Record.} OpenAI \texttt{gpt-5.5-2026-04-23}; manual rating 5; prompt asks
for a tall rolling toolbox with recessed wheels, a hinged front door, and a
telescoping rear handle.

\vspace{0.35em}
\textbf{Tools called.} \texttt{read\_file} $\times 11$,
\texttt{find\_examples} $\times 1$, \texttt{compile\_model} $\times 5$,
\texttt{probe\_model} $\times 5$, and \texttt{apply\_patch} $\times 7$.

\begin{itemize}
\setlength{\itemsep}{0.18em}
\setlength{\parskip}{0pt}
\setlength{\parsep}{0pt}
\item \textbf{Turns 1--5: workspace grounding.}
The agent reads \texttt{model.py} and SDK references for core types,
articulated objects, CadQuery, knobs and controls, and wheel/tire geometry.
\item \textbf{Turns 6--10: example retrieval and first construction.}
\texttt{find\_examples} retrieves wheel and tire patterns; the first patches
instantiate a toolbox body, recessed wheels, door articulation, and handle
mechanism.
\item \textbf{Turns 11--18: structured compile feedback.}
\texttt{compile\_model} returns typed failures, warnings, notes, and response
rules: invalid continuous-joint limits, floating wheels, handle overlap, and
disconnected-geometry warnings are surfaced as separate repair targets.
\item \textbf{Turns 19--23: targeted inspection.}
\texttt{probe\_model} snippets inspect AABBs, part summaries, and helper
availability. The agent uses these measurements and a lightweight
\texttt{catalog()} probe to select the exact-geometry helpers used in the next
repair.
\item \textbf{Turns 24--25: scoped repair and acceptance.}
\texttt{apply\_patch} adds rear guide bushings, restores exact visual names,
and scopes intentional overlaps for axle capture and telescoping rods. The
final \texttt{compile\_model} returns \texttt{status=success failures=0
warnings=0 notes=7}.
\end{itemize}
\vspace{0.1em}
\end{minipage}%
}

\vspace{0.6em}
\begin{lstlisting}[basicstyle=\ttfamily\fontsize{6}{6.5}\selectfont,breaklines=true,frame=single,columns=fullflexible]
<compile_signals>
<summary>
status=failure failures=1 warnings=0 notes=3
This is compile failure 4 in a row.
</summary>

<failures>
- [isolated_part] Floating disconnected component(s) detected.
part 'pull_handle' is disconnected from the grounded body rooted at 'body';
nearest_grounded_part='body'; approx_gap=0.006m.
</failures>

<notes>
- [allowed_overlap] allow_overlap('body','wheel_0'),
  elem_a='axle_stub_0', elem_b='rim':
  fixed axle stub is intentionally captured inside the wheel bore.
</notes>

<response_rules>
- Treat the compiler-owned floating/disconnected part finding as primary evidence
  before tuning authored exact checks.
- You are in a repair loop. A short probe_model snippet is likely to be more
  informative than another small placement or tolerance tweak.
</response_rules>
</compile_signals>
\end{lstlisting}

\vspace{0.4em}
\begin{lstlisting}[style=pythoncode,basicstyle=\ttfamily\fontsize{6}{6.5}\selectfont]
ctx.allow_overlap(
    body, wheel_0, elem_a="axle_stub_0", elem_b="rim",
    reason="The axle stub is intentionally captured inside the wheel bore.",
)

for guide_name in ("handle_guide_0", "handle_guide_1"):
    for rod_name in ("handle_rod_0", "handle_rod_1"):
        ctx.allow_overlap(
            body, pull_handle, elem_a=guide_name, elem_b=rod_name,
            reason="The rods intentionally pass through rear guide bushings.",
        )

rest_handle_pos = ctx.part_world_position(pull_handle)
with ctx.pose({handle_joint: HANDLE_TRAVEL}):
    extended_handle_pos = ctx.part_world_position(pull_handle)

ctx.check(
    "pull handle extends upward",
    rest_handle_pos is not None
    and extended_handle_pos is not None
    and extended_handle_pos[2] > rest_handle_pos[2] + 0.25,
)
\end{lstlisting}
\end{minipage}
\caption{Curated excerpt from a GPT-5.5 trace for a 5-star rolling toolbox asset.
The trace shows diverse tool use and highlights the structured feedback that the
harness provides to guide repair.}
\label{fig:appendix_trace_example}
\end{figure}
\clearpage

\paragraph{Context compaction.}
For long runs, the harness can compact older conversation history before the next model
call. Compaction is not performed every turn. It is triggered either by hard context-window
pressure or by a soft repair-plateau rule: repeated compile failures, sufficient context
pressure, and enough compactable history. The policy preserves the immutable run prefix
and the most recent raw tail, while replacing older intermediate history with a compact
summary of task requirements, constraints, tool findings, compile state, and next steps.
OpenAI uses the Responses API compaction endpoint; Gemini uses a separate JSON-summary
prompt. Anthropic runs do not use provider-side compaction in the current implementation.

\begin{table}[h]
\centering
\footnotesize
\setlength{\tabcolsep}{4pt}
\caption{Context compaction thresholds used by the current harness. Hard compaction fires
at $0.9$ times the listed pressure threshold. Soft compaction can fire earlier during
repeated compile-failure plateaus.}
\label{tab:appendix_compaction}
\begin{tabular}{p{0.28\linewidth} p{0.22\linewidth} p{0.42\linewidth}}
\toprule
\textbf{Provider/model family} & \textbf{Pressure threshold} & \textbf{Compaction mechanism} \\
\midrule
OpenAI GPT-5.4/5.5~\citep{openai2026gpt54,openai2026gpt55} & 272k prompt tokens & Responses API compaction over older input items. \\
OpenAI GPT-5.2 and GPT-5.2/5.3-Codex~\citep{openai2025gpt52,openai2025gpt52codex,openai2026gpt53codex} & 280k prompt tokens & Responses API compaction over older input items. \\
Gemini 3.1 Pro~\citep{google2026gemini31pro} / Gemini 2.5 & 700k prompt tokens & JSON summary produced by a Gemini compaction prompt. \\
Anthropic & -- & No provider-side compaction in the current implementation. \\
\bottomrule
\end{tabular}
\end{table}

\section{Additional Related Work}
\paragraph{Reconstructing articulated objects.}

Several prior works considered the problem of reconstructing articulated objects.
Shape2Motion~\cite{wang19shape2motion:} starts from a 3D point cloud and segments it in parts and their joints.
The work of~\cite{li20category-level} uses canonical spaces to solve a similar problem and CAPTRA~\cite{weng21captra:} further tracks the motion of parts over time.
A-SDF~\cite{mu21a-sdf:} introduce an articulated version of sign distance functions to model articulated objects.
DITTO~\cite{jiang22ditto:} reconstructs articulated 3D objects from a pair of images showing different poses and PARIS~\cite{liu2023paris} does so in a self-supervised manner.
Differently from these works, our goal is to generate a new articulated object from a textual prompt.

\section{Societal Impacts}

\agentname{} can have positive societal impacts by reducing the cost of producing
articulated 3D assets for animation, games, education, robotic simulation, and
embodied-AI research.
The resulting assets may help researchers build more diverse simulation environments
and study manipulation, planning, and interaction without manually authoring every
object.
At the same time, scalable asset generation can be misused to create synthetic
environments or objects for deceptive visual content, unauthorized replication of
proprietary designs, or unsafe embodied-agent training.
Generated geometry and articulation may also contain errors that propagate to
downstream simulators or robots, especially in safety-critical manipulation settings.
We therefore view \agentname{} primarily as a research tool, and recommend that
practical deployments validate generated assets in the target domain, respect object
and dataset licensing constraints, and apply appropriate access controls or monitoring
when generated assets are used in higher-risk settings.

%% file: tables/backend_retention.tex
\begin{table}[h]
\centering
\footnotesize
\caption{Manual curation retention by backend. Retained assets are those rated 4 or 5.}
\begin{tabular}{lrrr}
\toprule
\textbf{Backend} & \textbf{Generated} & \textbf{Retained} & \textbf{Retention rate} \\
\midrule
GPT-5.4 & 6,601 & 5,903 & 89.4\% \\
GPT-5.5 & 4,010 & 3,828 & 95.5\% \\
GPT-5.4/GPT-5.5 total & 10,611 & 9,731 & 91.7\% \\
Gemini 3.1 Pro & 298 & 287 & 96.3\% \\
\bottomrule
\end{tabular}
\label{tab:backend_retention}
\end{table}

%% file: figures/supp/construction_style.tex
\begin{figure*}[t]
    \centering
    \begin{subfigure}[t]{0.49\textwidth}
        \centering
        \includegraphics[width=\linewidth]{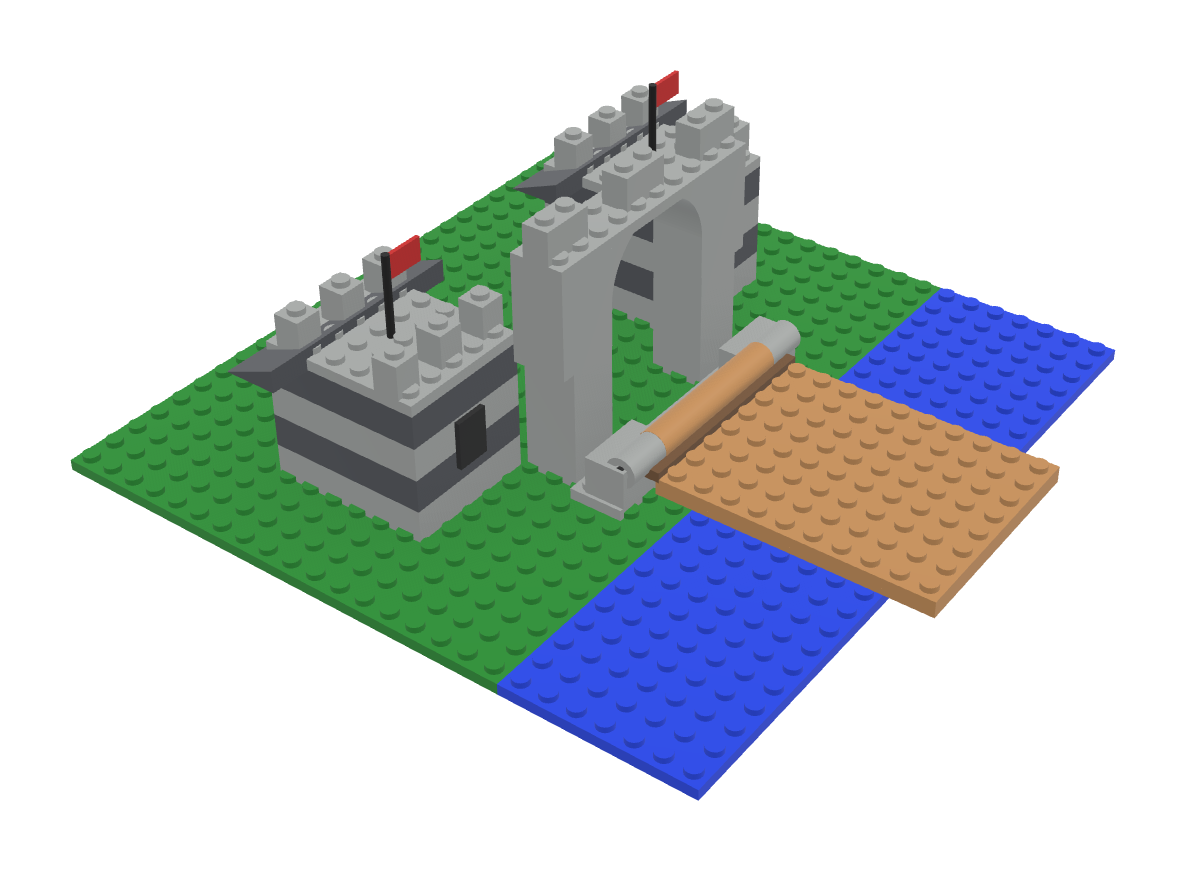}
        \caption{Drawbridge lowered.}
    \end{subfigure}
    \hfill
    \begin{subfigure}[t]{0.49\textwidth}
        \centering
        \includegraphics[width=\linewidth]{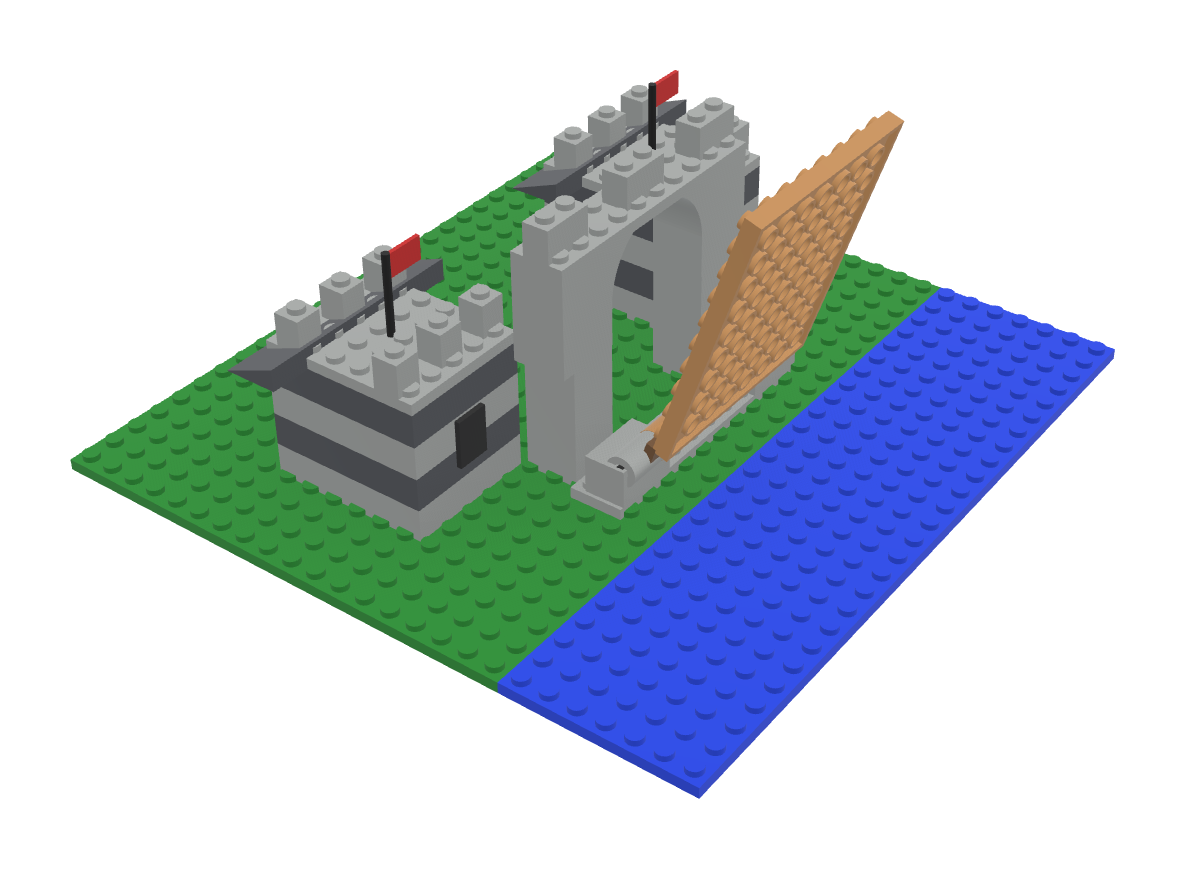}
        \caption{Drawbridge raised.}
    \end{subfigure}
    \vspace{-0.5em}
    \caption{Example of the \agentname{} SDK's expressivity and style controllability.
    The same block-based gatehouse is rendered in two drawbridge states, showing that the generated asset controls both construction idiom and articulation.}
    \label{fig:construction_style_gatehouse}
\end{figure*}

%% file: figures/supp/dataset_stats.tex
\begin{figure}[t]
\centering
    \begin{subfigure}{0.59\textwidth}
    \centering
    \includegraphics[trim={0 10bp 0bp 0}, clip, width=\linewidth]{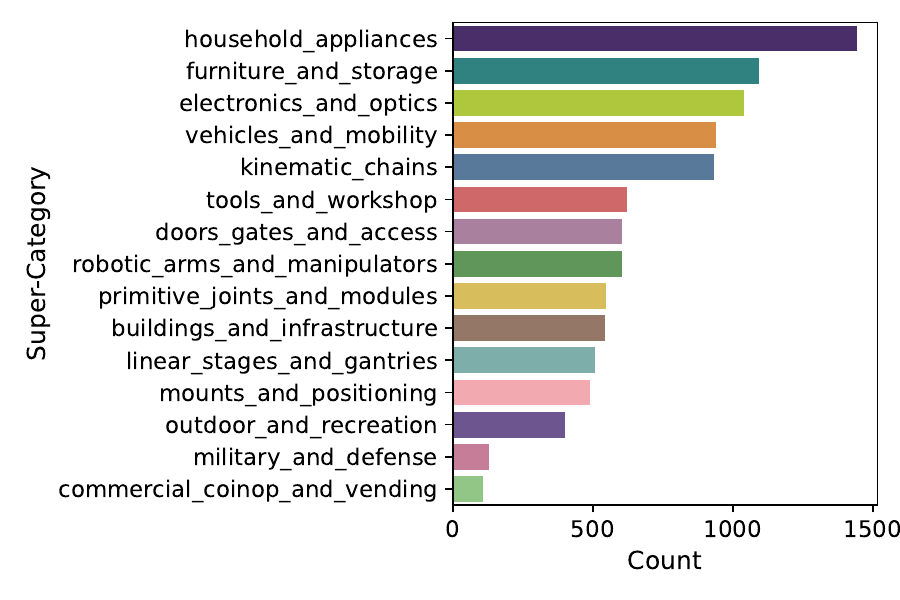}
    \caption{Object count by super-categories.}
    \label{fig:super_cat_hist}
    \end{subfigure}
    \hfill
    \begin{subfigure}{0.39\textwidth}
    \centering
    \includegraphics[trim={0 10bp 0bp 0}, clip, width=\linewidth]{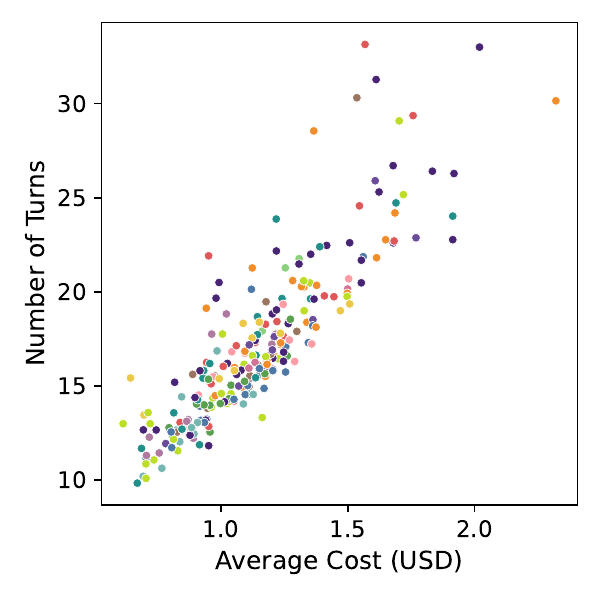}
    \caption{Average cost and turns per category.}
    \label{fig:category_cost_scatter}
    \end{subfigure}
\caption{Statistics of \datasetname{}: (a) Number of objects in each super-category. (b) Average cost and number of turns for each object category, color-coded by super-category.}
\label{fig:data_statistics}
\end{figure}

\begin{figure}[t]
\centering
\includegraphics[width=0.85\linewidth]{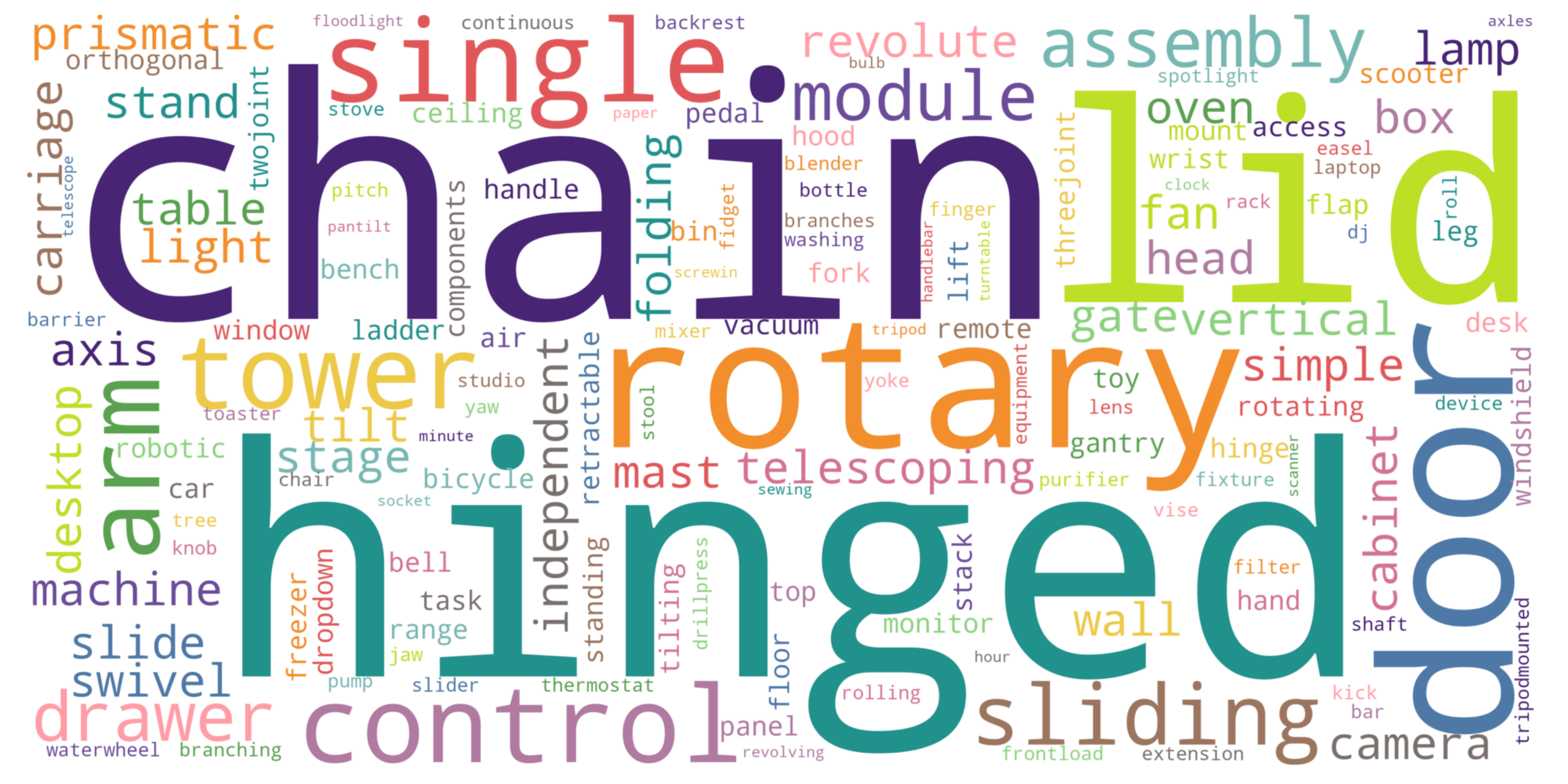}
\caption{Word cloud of object names in \datasetname{}.}
\label{fig:object_wordcloud}
\end{figure}

%% file: figures/supp/dataset_stats_dist.tex
\begin{figure}[t]
\centering
\includegraphics[width=\linewidth]{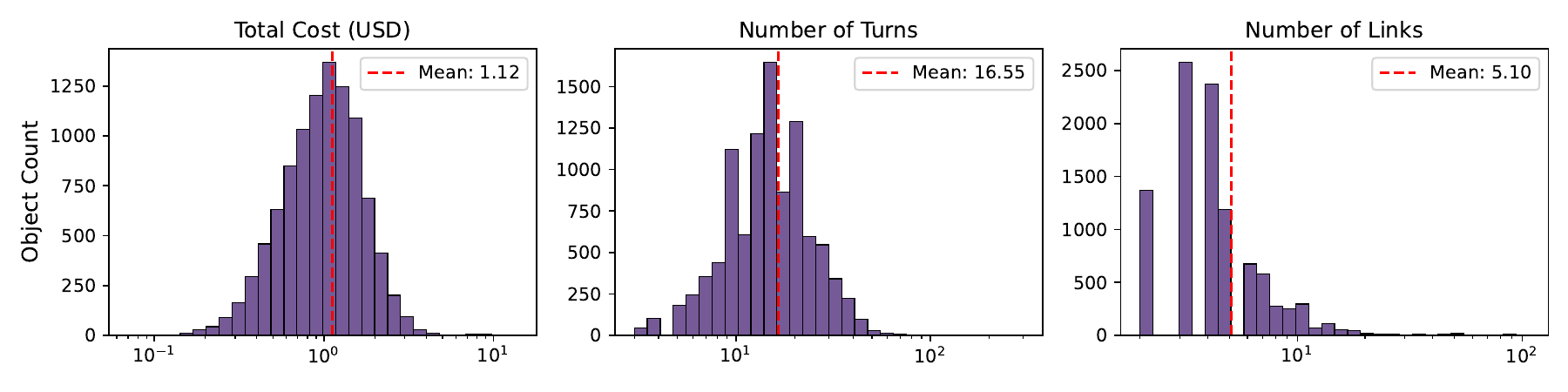}
\vspace{-2em}
\caption{Distribution of cost, number of turns, and number of links per object in \datasetname{}.}%
\label{fig:costs_turns_link_histogram}
\end{figure}

\begin{figure}[t]
\centering
\includegraphics[width=\linewidth]{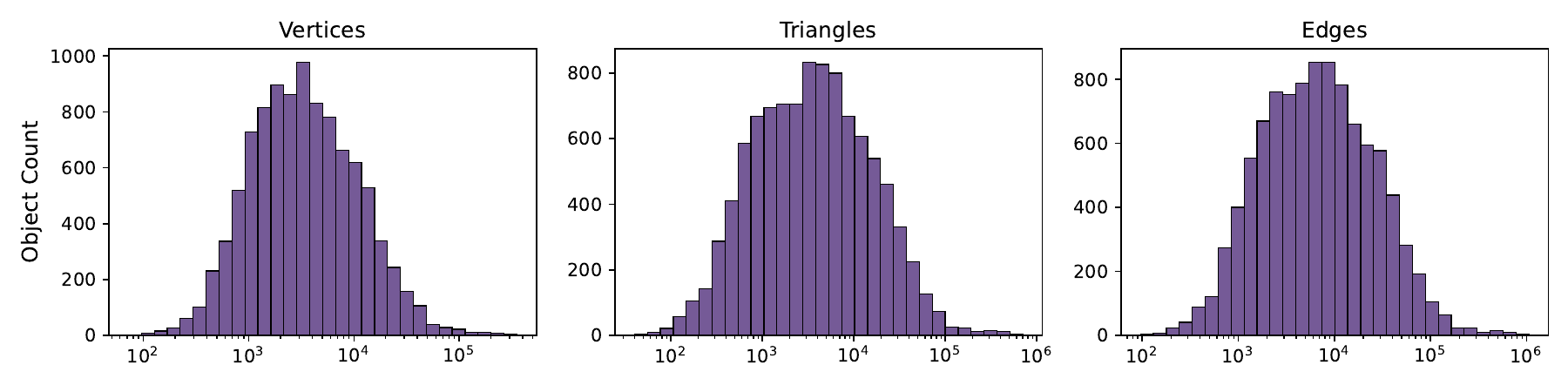}
\vspace{-2em}
\caption{Mesh statistics of \datasetname{}.}
\label{fig:mesh_distributions}
\end{figure}

\begin{figure}[t]
\centering
\includegraphics[width=\linewidth]{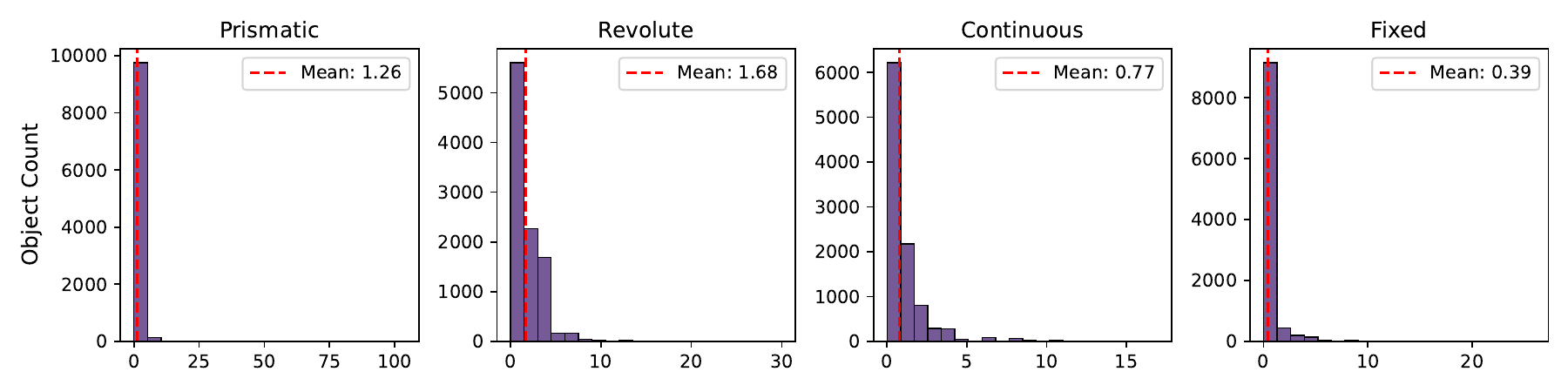}
\vspace{-2em}
\caption{Distribution of URDF joints per object in \datasetname{}.}
\label{fig:joint_distributions}
\end{figure}

%% file: tables/api_costs.tex
\begin{table}[t]
\centering
\footnotesize
\setlength{\tabcolsep}{3pt}
\caption{API cost and turn statistics for the main generation backends. Costs are in USD
and are computed from available per-record cost logs.}
\begin{tabular}{lrrrrr}
\toprule
\textbf{Backend} & \textbf{Cost logs} & \textbf{Retained} & \textbf{Total cost} &
\textbf{Mean/med. cost} & \textbf{Mean/med. turns} \\
\midrule
GPT-5.4 & 6,572 & 5,903 & \$6,693.89 & \$1.019 / \$0.887 & 16.9 / 15 \\
GPT-5.5 & 4,010 & 3,828 & \$5,305.18 & \$1.323 / \$1.240 & 16.4 / 15 \\
Gemini 3.1 Pro & 298 & 287 & \$387.91 & \$1.302 / \$0.711 & 19.5 / 10 \\
\midrule
Total & 10,880 & 10,018 & \$12,386.99 & \$1.139 / \$1.030 & 16.8 / 15 \\
\bottomrule
\end{tabular}
\label{tab:api_costs}
\end{table}

%% file: figures/supp/image_conditioning.tex
\begin{figure*}[t]
    \centering
    \includegraphics[width=\textwidth,keepaspectratio]{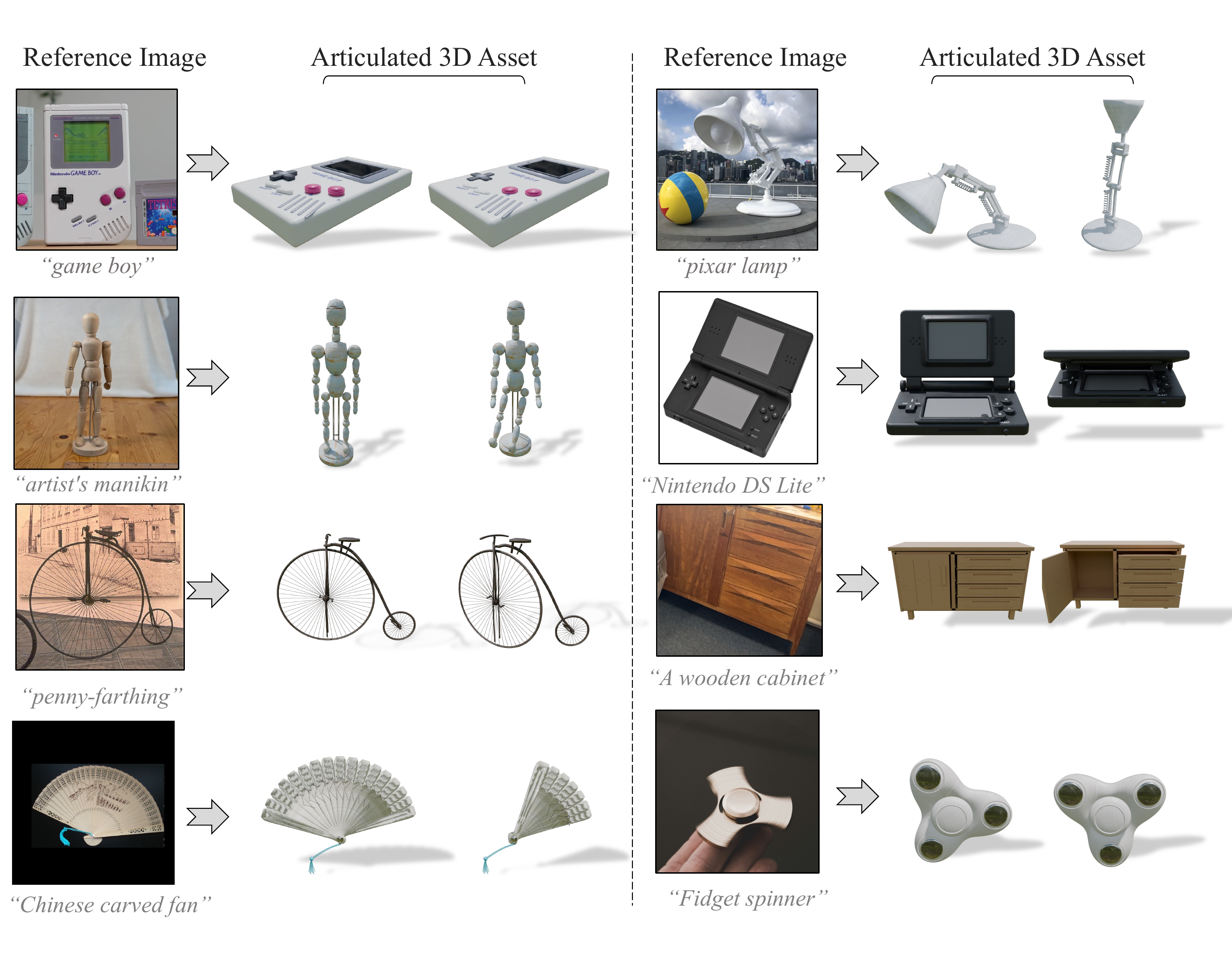}
    \vspace{-2em}
    \caption{More examples of image-conditioned articulated 3D asset generations produced by \agentname{}.}
    \label{fig:image_conditioning_supp}
\end{figure*}

%% file: figures/supp/scene_recon.tex
\begin{figure*}[t]
    \centering
    \includegraphics[width=\textwidth,keepaspectratio]{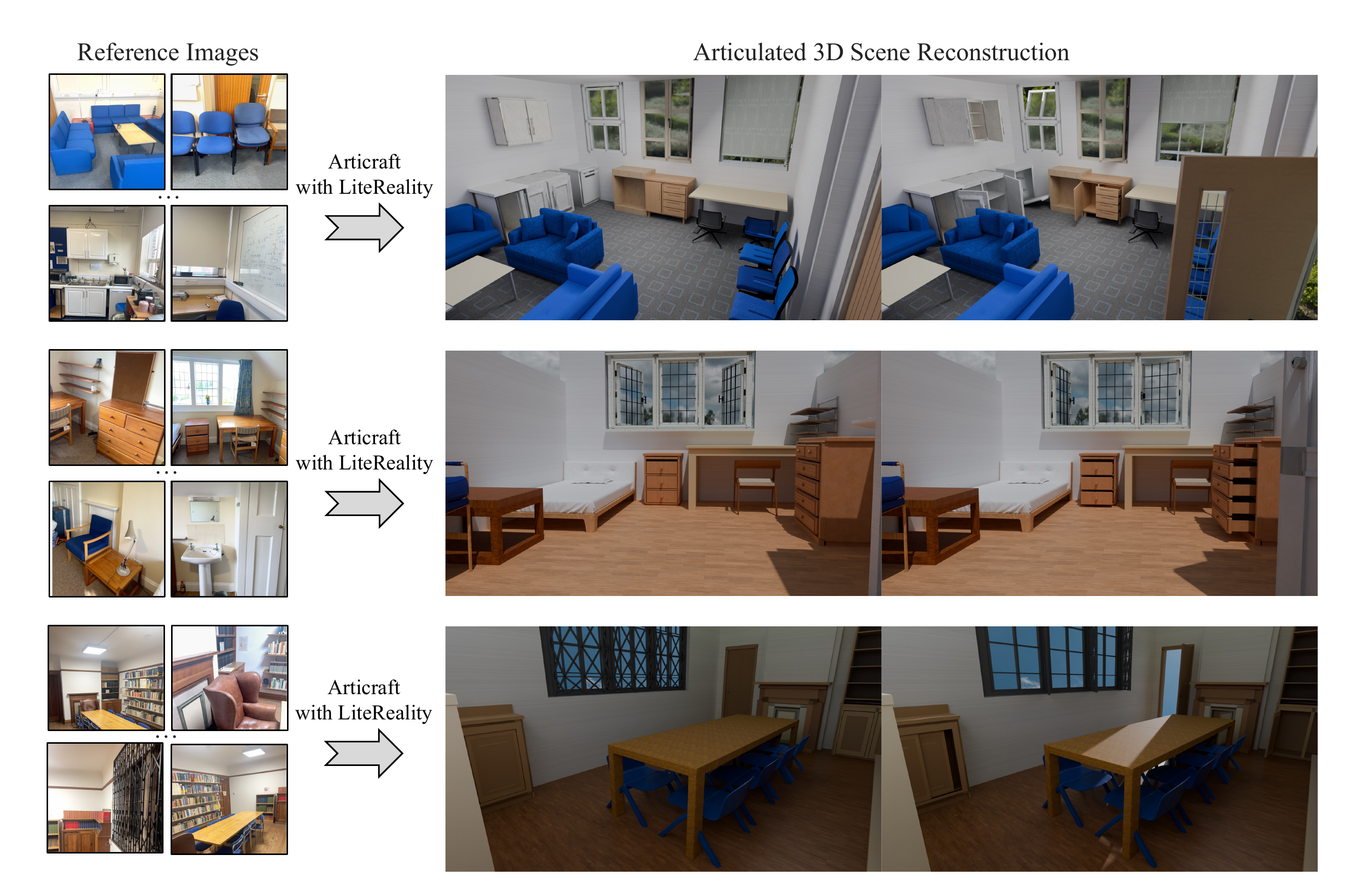}
    \caption{\agentname{} integrates seamlessly into the LiteReality \cite{huang2025literealitygraphicsready3dscene} pipeline: in place of the original retrieval-based object stage, \agentname{} generates articulated assets on the fly, which are well-aligned with the reconstructed scene context.}
    \label{fig:scene_recon_supp}
\end{figure*}

%% file: figures/supp/failure_caes.tex
\begin{figure*}[h]
    \centering
    \includegraphics[trim={20bp 10bp 40bp 10bp}, clip, width=\textwidth,keepaspectratio]{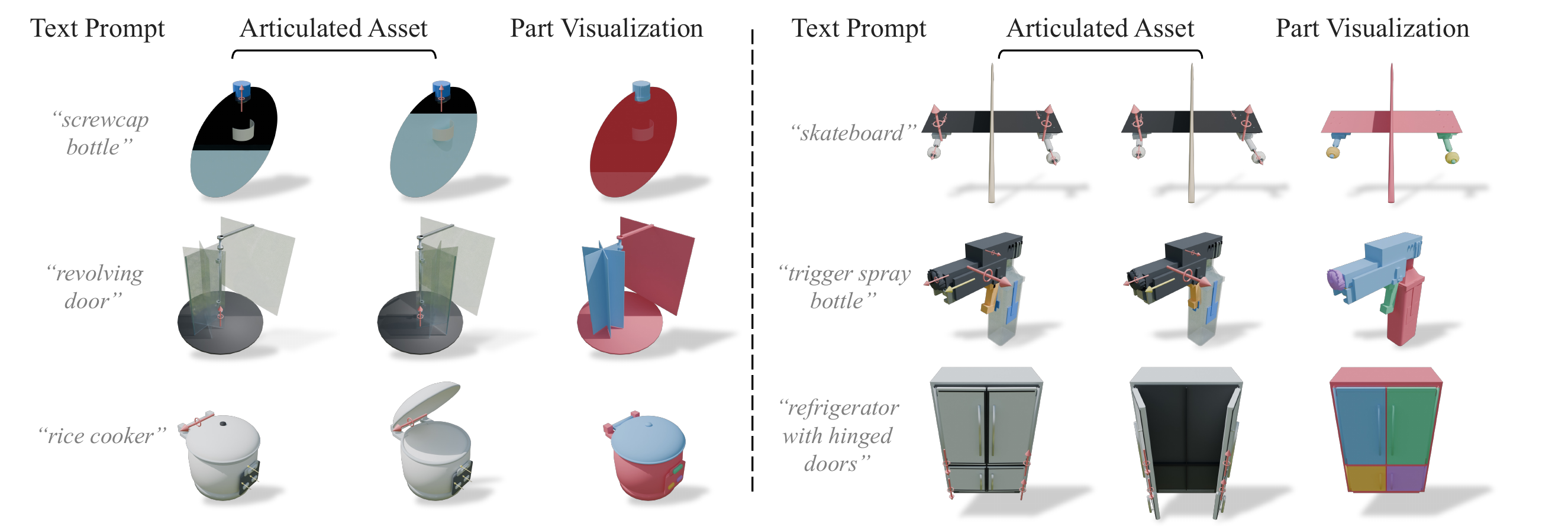}
    \caption{Examples of failure cases of \agentname{}.}
    \label{fig:failure_cases}
\end{figure*}

%% file: figures/supp/results_supp.tex
\begin{figure*}[t]
    \centering
    \includegraphics[trim={20bp 10bp 40bp 10bp}, clip, width=\textwidth,keepaspectratio]{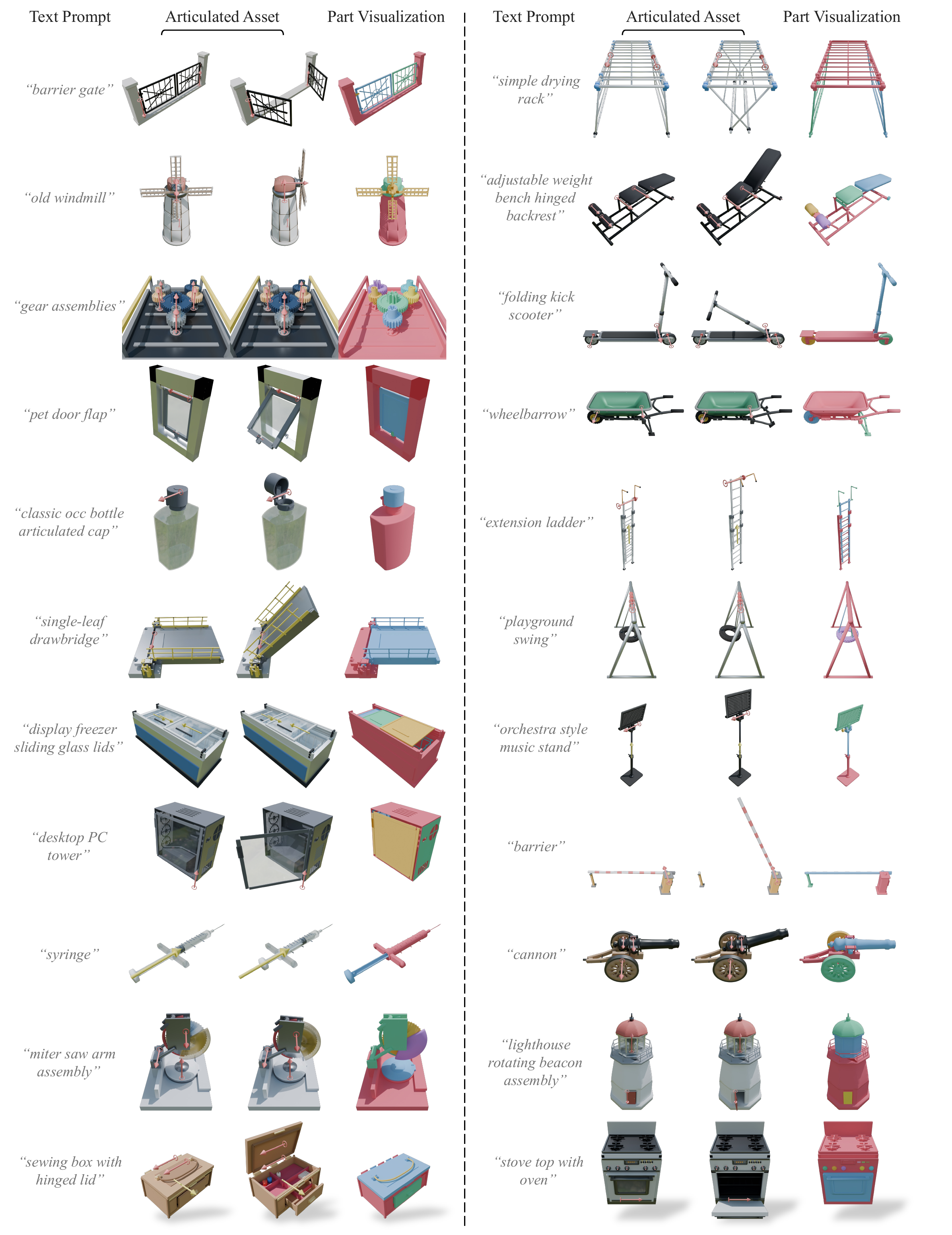}
    \caption{Additional articulated assets generated by \agentname{}. The results span a broad range of object categories and scales, from small everyday items to mechanical assemblies, furniture, appliances, and large-scale outdoor structures.}
    \label{fig:artcraft_results_supp}
\end{figure*}

%% file: tables/particulate-category-result.tex
\begin{table*}[t]
    \centering
    \footnotesize
    \setlength{\tabcolsep}{1pt}
    \renewcommand{\arraystretch}{1.03}
    \caption{
        Per-category evaluation results comparing Particulate and Particulate-\agentname{} on the 14 categories in the Lightwheel benchmark.
        Categories shaded in \colorbox{green!15}{green} are absent from the training data of Particulate.
        The better values are colored in \best{maroon}.
        The \datasetname{} dataset brings more significant improvements on previously out-of-distribution categories.
    }%
    \label{tab:particulate_category_result}
    \newcommand{\catcell}[2]{\parbox[c]{#1}{\centering #2}}
    \resizebox{\linewidth}{!}{%
    \begin{tabular}{@{}llcccccc>{\columncolor{green!15}}ccc>{\columncolor{green!15}}c>{\columncolor{green!15}}c>{\columncolor{green!15}}ccc@{}}
        \toprule
        \multicolumn{2}{c}{} &
        \multicolumn{14}{c}{\textbf{Categories}} \\
        \cmidrule(lr){3-16}
        \textbf{Model} & \textbf{Metric} &
        \catcell{1cm}{Blender} &
        \catcell{1cm}{Coffee\\Machine} &
        \catcell{1cm}{Dish-\\washer} &
        \catcell{1cm}{Electric\\Kettle} &
        \catcell{1cm}{Micro-\\wave} &
        \catcell{1cm}{Oven} &
        \catcell{1cm}{Range\\Hood} &
        \catcell{1cm}{Refri-\\gerator} &
        \catcell{1cm}{Sink} &
        \catcell{1cm}{Stand\\Mixer} &
        \catcell{1cm}{Stove} &
        \catcell{1cm}{Stovetop} &
        \catcell{1cm}{Toaster} &
        \catcell{1cm}{Toaster\\Oven} \\
        \midrule
        Particulate~\citep{li2026particulate}
        & mIoU (rest) & \best{0.708} & 0.435 & 0.811 & \best{0.674} & 0.802 & 0.521 & 0.303 & 0.633 & \best{0.508} & 0.284 & 0.543 & 0.373 & 0.509 & 0.490 \\
        & gIoU (rest) & \best{0.501} & 0.038 & 0.689 & \best{0.390} & 0.632 & 0.203 & -0.088 & 0.498 & \best{0.425} & -0.071 & 0.292 & -0.007 & 0.199 & 0.134 \\
        & PC (rest)   & 0.121 & 0.278 & 0.088 & \best{0.207} & 0.115 & 0.234 & 0.273 & 0.072 & \best{0.047} & 0.281 & 0.141 & 0.195 & 0.214 & 0.252 \\
        & gIoU (art.) & \best{0.456} & 0.033 & 0.640 & \best{0.330} & 0.580 & 0.179 & -0.088 & 0.452 & \best{0.423} & -0.083 & 0.274 & -0.012 & 0.181 & 0.110 \\
        & PC (art.)   & \best{0.171} & 0.321 & 0.177 & \best{0.276} & 0.116 & 0.242 & 0.276 & 0.075 & 0.158 & 0.482 & 0.147 & \best{0.203} & 0.220 & 0.254 \\
        & OC (art.)   & \best{0.019} & 0.012 & \best{0.027} & 0.018 & 0.000 & \best{0.005} & 0.002 & \best{0.001} & \best{0.003} & 0.038 & \best{0.001} & 0.001 & 0.001 & 0.002 \\
        \midrule
        \textbf{Particulate-\agentname{}}
        & mIoU (rest) & 0.546 & \best{0.457} & \best{0.820} & 0.666 & \best{0.812} & \best{0.645} & \best{0.512} & \best{0.699} & 0.436 & \best{0.441} & \best{0.692} & \best{0.506} & \best{0.580} & \best{0.650} \\
        & gIoU (rest) & 0.360 & \best{0.070} & \best{0.704} & 0.374 & \best{0.658} & \best{0.435} & \best{0.232} & \best{0.577} & 0.289 & \best{0.179} & \best{0.567} & \best{0.225} & \best{0.383} & \best{0.397} \\
        & PC (rest)   & 0.121 & \best{0.261} & \best{0.083} & 0.213 & \best{0.107} & \best{0.167} & \best{0.174} & \best{0.066} & 0.091 & \best{0.196} & \best{0.084} & \best{0.158} & \best{0.136} & \best{0.177} \\
        & gIoU (art.) & 0.331 & \best{0.065} & \best{0.652} & 0.313 & \best{0.608} & \best{0.407} & \best{0.214} & \best{0.526} & 0.286 & \best{0.146} & \best{0.518} & \best{0.214} & \best{0.356} & \best{0.347} \\
        & PC (art.)   & 0.181 & \best{0.311} & \best{0.175} & 0.277 & \best{0.108} & \best{0.185} & \best{0.176} & \best{0.074} & \best{0.112} & \best{0.338} & \best{0.098} & 0.238 & \best{0.178} & \best{0.177} \\
        & OC (art.)   & 0.020 & \best{0.009} & 0.029 & 0.018 & 0.000 & 0.010 & \best{0.001} & 0.002 & 0.005 & \best{0.034} & 0.002 & \best{0.000} & 0.001 & 0.002 \\
        \bottomrule
    \end{tabular}%
    }
\end{table*}

%% file: tables/appendix_ablation_stats.tex
\begin{table}[t]
\centering
\footnotesize
\setlength{\tabcolsep}{2.5pt}
\caption{Run statistics for the LLM and reasoning-effort ablation in \cref{fig:ablation}. Token counts are rounded prompt/output tokens from the recorded provider logs; visual element counts are measured from the materialized URDF files.}
\begin{tabular}{llccccc}
\toprule
\textbf{Provider} & \textbf{Model} & \textbf{Effort} & \textbf{Turns} & \textbf{Cost} & \textbf{Visuals} & \textbf{Prompt/out. tokens} \\
\midrule
OpenAI & \texttt{gpt-5.5-2026-04-23} & low & 17 & \$0.60 & 39 & 362K / 4.8K \\
OpenAI & \texttt{gpt-5.5-2026-04-23} & med & 15 & \$1.08 & 51 & 398K / 11.1K \\
OpenAI & \texttt{gpt-5.5-2026-04-23} & high & 22 & \$1.37 & 78 & 961K / 19.2K \\
Google & \texttt{gemini-3.1-pro-preview} & high & 26 & \$3.14 & 13 & 1.54M / 5.3K \\
Anthropic & \texttt{claude-opus-4-7} & high & 26 & \$1.97 & 43 & 1.61M / 27.6K \\
\bottomrule
\end{tabular}
\label{tab:appendix_ablation_stats}
\end{table}